\documentclass[sigconf]{acmart}

\AtBeginDocument{%
  }

\copyrightyear{2026}
\acmYear{2026}
\setcopyright{cc}
\setcctype{by}
\acmConference[WWW '26]{Proceedings of the ACM Web Conference 2026}{April 13--17, 2026}{Dubai, United Arab Emirates}
\acmBooktitle{Proceedings of the ACM Web Conference 2026 (WWW '26), April 13--17, 2026, Dubai, United Arab Emirates}
\acmPrice{}
\acmDOI{10.1145/3774904.3792979}
\acmISBN{979-8-4007-2307-0/2026/04}

\usepackage{subcaption}
\usepackage{graphicx}
\usepackage{algorithm}
\usepackage{graphicx}
\usepackage{enumitem}
\usepackage{xcolor}
\usepackage{multirow}
\usepackage{multicol}
\usepackage{makecell}
\usepackage{booktabs}
\usepackage{float}
\usepackage{algpseudocode}
\usepackage{color}

\algnewcommand{\Input}{\item[\textbf{Input:}]}
\algnewcommand{\Output}{\item[\textbf{Output:}]}

\begin{document}

\title{Bridging Cognitive Neuroscience and Graph Intelligence: \\Hippocampus-Inspired Multi-View Hypergraph Learning for Web Finance Fraud}

\author{Rongkun Cui}
  \orcid{0009-0000-9879-8775}
\affiliation{%
  \institution{School of Computer Science and Technology, Tongji University}
  \city{Shanghai}
  \country{China}
}
\email{jokercui@tongji.edu.cn}

\author{Nana Zhang}
\orcid{0000-0002-9843-5269}
\affiliation{%
  \institution{School of Computer Science and Technology, Donghua University}
  \city{Shanghai}
  \country{China}
}
\email{nnzhang@dhu.edu.cn}

\author{Kun Zhu}
\authornote{Corresponding author.}
\orcid{0000-0002-5773-5089}
\affiliation{%
  \institution{School of Computer Science and Technology, Tongji University}
  \city{Shanghai}
  \country{China}
}
\email{kzhu00@tongji.edu.cn}

\author{Qi Zhang}
\orcid{0000-0002-1037-1361}
\affiliation{%
  \institution{School of Computer Science and Technology, Tongji University}
  \city{Shanghai}
  \country{China}
}
\email{zhangqi\_cs@tongji.edu.cn}

\renewcommand{\shortauthors}{Rongkun Cui, Nana Zhang, Kun Zhu, and Qi Zhang}

\begin{abstract}
Online financial services constitute an essential component of contemporary web ecosystems, yet their openness introduces substantial exposure to fraud that harms vulnerable users and weakens trust in digital finance. Such threats have become a significant web harm that erodes societal fairness and affects the well-being of online communities. However, existing detection methods based on graph neural networks (GNNs) struggle with two persistent challenges: (1) long-tailed data distributions, which obscure rare but critical fraudulent cases, and (2) fraud camouflage, where malicious transactions mimic benign behaviors to evade detection. To fill these gaps, we propose HIMVH, a Hippocampus-Inspired Multi-View Hypergraph learning model for web finance fraud detection. Specifically, drawing inspiration from the scene conflict monitoring role of the hippocampus, we design a cross-view inconsistency perception module that captures subtle discrepancies and behavioral heterogeneity across multiple transaction views. This module enables the model to identify subtle cross-view conflicts for detecting online camouflaged fraudulent behaviors. Furthermore, inspired by the match-mismatch novelty detection mechanism of the CA1 region, we introduce a novelty-aware hypergraph learning module that measures feature deviations from neighborhood expectations and adaptively reweights messages, thereby enhancing sensitivity to online rare fraud patterns in the long-tailed settings. Extensive experiments on six web-based financial fraud datasets demonstrate that HIMVH achieves 6.42\% improvement in AUC, 9.74\% in F1 and 39.14\% in AP on average over 15 SOTA models.
\end{abstract}




\maketitle

\section{Introduction}

Web finance has become a critical part of digital life and provides essential channels for credit access and fund movement across web communities. This shift creates new forms of exposure to financial harm, as web transaction networks allow fraudulent behavior to propagate across services and reach large populations. This problem now stands as a major web harm with real societal consequences, since web finance fraud reduces trust in digital public infrastructure, erodes societal fairness by intensifying the disadvantages faced by groups with restricted financial capacity, and intensifies existing economic vulnerability \cite{Strelcenia2023,zhang2024dig,aftabi2023fraud}. As financial interactions across the web grow more complex and interlinked, the need for reliable intelligence systems that counter online fraud becomes a central requirement for a safe and fair society.

\begin{figure}[t]
    \centering

    \begin{subfigure}{0.2\textwidth}
        \centering
        \includegraphics[width=\linewidth]{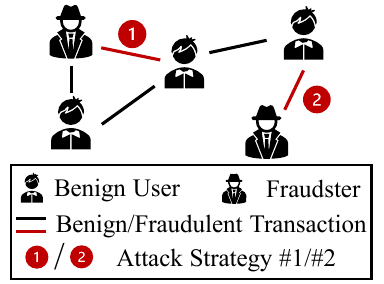}
        \caption{Two attack strategies}
        \label{fig:two-attack}
    \end{subfigure}
    \hfill
    \begin{subfigure}{0.235\textwidth}
        \centering
        \includegraphics[width=\linewidth]{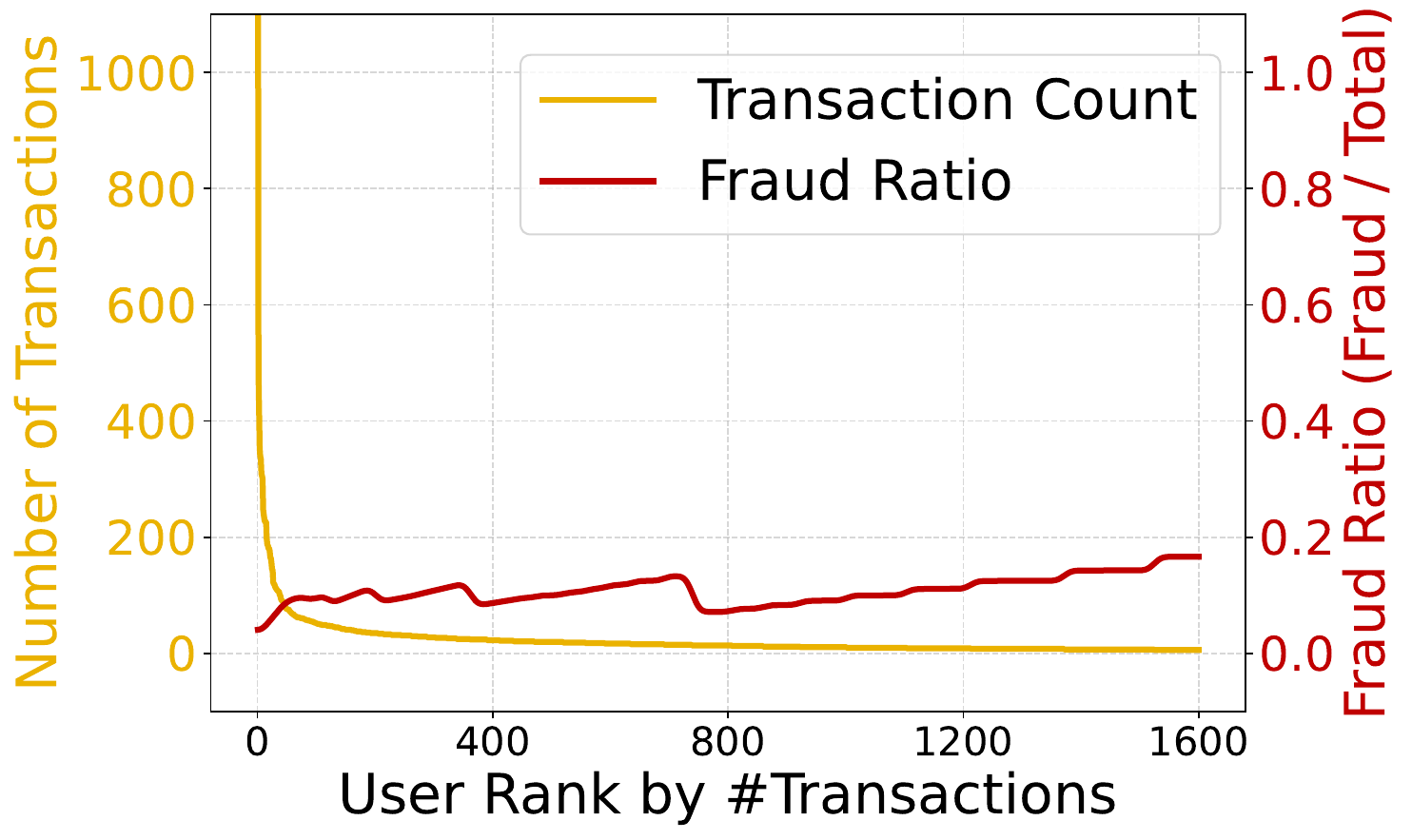}
        \caption{Fraud distribution on S-FFSD}
        \label{fig:fraud-dist}
    \end{subfigure}

    \caption{(a) Two attack strategies targeting the head and tail of the graph.
    (b) Fraud distribution on the S-FFSD dataset.}
    \label{fig:figure1}
\end{figure}

In response to the substantial societal impact of web finance fraud, extensive research \cite{Belle2023,Cheng2023anti,Li2020Flow} has been conducted to develop models capable of accurately identifying online fraudulent transactions. Initial methods used rule-based heuristics but were gradually replaced by traditional machine learning (e.g., decision trees) \cite{zou2025effective,nti2022scalable}. As large-scale online transactions grow, deep learning techniques gain more popularity due to their capability to capture complex patterns \cite{Wang2017Session,Zhuojia2025}. With the growing sophistication of online fraudulent behaviors \cite{Shi2023Cost}, graph neural networks (GNNs) have become a mainstream paradigm in web finance fraud detection due to their capability to capture and learn from the intricate relational structures embedded within web-based financial systems \cite{huang2022auc,lakhan2022its}.

Nevertheless, these web finance fraud detection methods based on GNNs still suffer from two significant limitations: (1) \textbf{Reduced detectability of fraudulent instances in the distribution tail}. Fraud in the tail of the distribution is rare and atypical, with limited representative samples. This scarcity hinders effective learning and causes models to overlook subtle yet critical fraud \cite{han2025mitigating}. As illustrated in Figure 1 (a), attack strategy \#2 targets sparse regions of the transaction graph, which correspond to the tail of data distribution. With limited structural context, these regions are more vulnerable to undetected attacks \cite{Luo2024MLa}. As shown in Figure 1 (b), the distribution in the S-FFSD dataset reveals a higher fraud ratio in tail regions, and highlights the greater risk in low-density areas. (2) \textbf{Imitation-based camouflaged fraud}. These fraudulent transactions deliberately imitate legitimate behaviors and thus become difficult to distinguish with conventional graph patterns \cite{haghighi2024tropical,xiao2024vecaug, Ou2025}.  The aforementioned challenges remain inadequately addressed. In this context, brain-inspired intelligence emerges as a compelling alternative, offering a biologically grounded foundation for contextual understanding, adaptive learning, and resilience against uncertainty. These capabilities are crucial for detecting sophisticated fraudulent behaviors in web-based financial systems.

\begin{figure}[t]
  \centering
    \begin{subfigure}{0.42\textwidth}
        \centering
        \includegraphics[width=\linewidth]{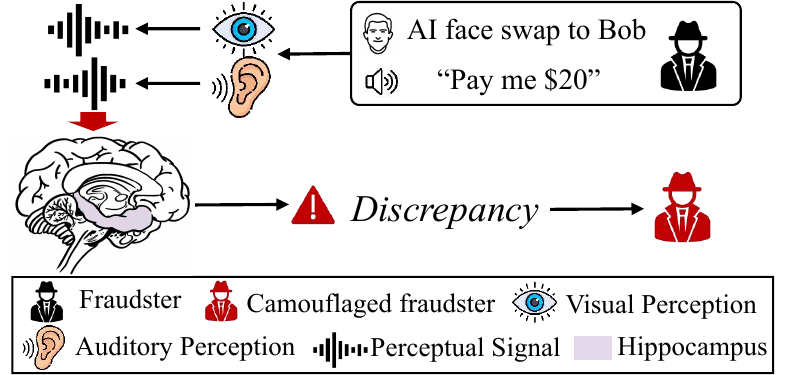}
        \caption{Scene conflict detection mechanism in Hippocampus}
        \label{fig:two-attack}
    \end{subfigure}

    \begin{subfigure}{0.45\textwidth}
        \centering
        \includegraphics[width=\linewidth]{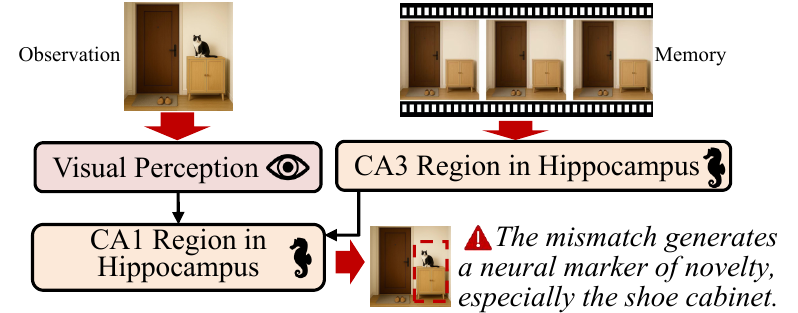}
        \caption{Novelty Detection Mechanism in Hippocampus}
        \label{fig:two-attack}
    \end{subfigure}
  \caption{Two mechanisms in hippocampus.}
  \label{Figure3}
\end{figure}

Motivated by the aforementioned brain-inspired intelligence, we propose HIMVH, a \textbf{H}ippocampus-\textbf{I}nspired \textbf{M}ulti-\textbf{V}iew \textbf{H}ypergraph learning model to detect web finance fraud. Figure 2 presents the two hippocampal mechanisms that serve as the foundation of our model, which will be elaborated in detail in the Preliminaries section. To capture high-order interactions in complex online transactions, we build a multi-view hypergraph where each hyperedge encodes context-aware group relations across views. Based on this, a hippocampal discrepancy perception mechanism is designed to identify latent cross-view inconsistencies in camouflaged fraud. Furthermore, a novelty-aware hypergraph neural network, inspired by the match–mismatch novelty detection in CA1 region, estimates prediction deviations within local neighborhoods and amplifies fraud responses to detect rare fraud patterns in long-tailed distributions. The main contributions can be summarized as follows:
\begin{itemize}
\item We propose HIMVH, the first work to reformulate the message-passing mechanism in graph learning through a brain-inspired perspective. We develop a CA1-inspired novelty-aware hypergraph network that quantifies local deviations to adaptively reweight messages, and enhance the representation of sparse minority samples under long-tailed distributions.
\item We design a hippocampal cross-view discrepancy perception module that systematically captures latent conflicts and heterogeneity of camouflaged online transactions across multiple views. 
\item Experimental results on six web finance fraud datasets underscore the HIMVH’s superiority, which achieves an average improvement of 6.42\% in AUC, 9.74\% in F1 score, and 39.14\% in AP compared to 15 SOTA models.
\end{itemize}

\begin{figure*}[htbp]
\centering
\includegraphics[width=0.95\textwidth]{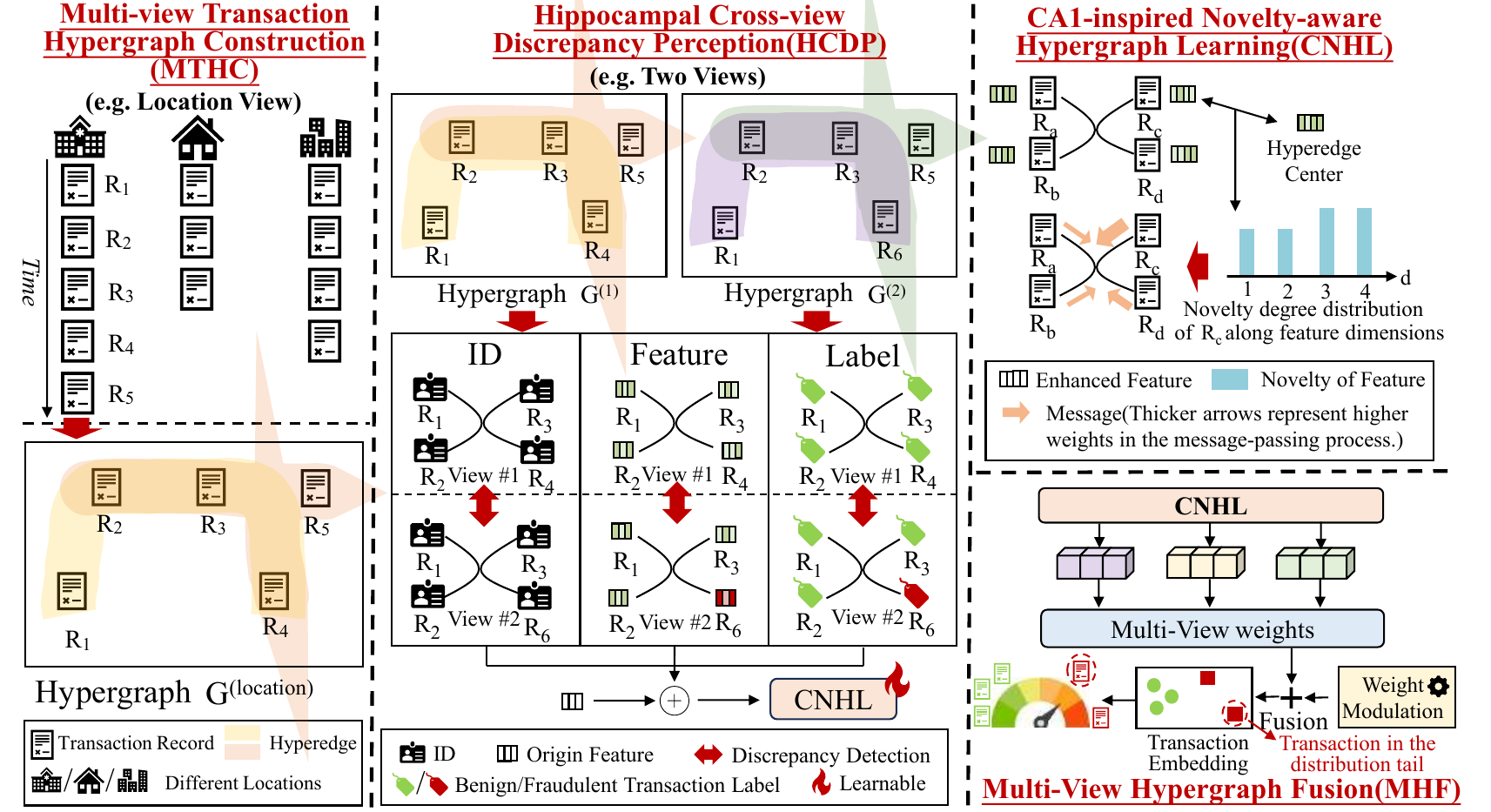} 
\caption{Overview of the proposed HIMVH model.}
\label{fig3}
\end{figure*}

\section{Related Work}
This section reviews existing studies from two aspects, namely general web finance fraud detection methods and graph-based financial risk control. These perspectives delineate the advances and constraints of current research and underscore the need for more robust graph-driven solutions that strengthen the integrity and trustworthiness of web finance ecosystems.
\subsection{Web Finance Fraud Detection}
During the initial stages of web finance fraud detection, traditional machine learning techniques were predominantly employed \cite{singh2022financial, sahin2011detecting}. Ensemble methods such as GBT \cite{xia2023credit,zhao2018xgbod,tang2023gadbench} provide competitive accuracy by aggregating weak learners, and allow them to handle large-scale, highly imbalanced datasets while maintaining robustness to noisy transactional patterns.

The emergence of deep learning has led to a paradigm shift in web finance fraud detection. Conventional machine learning techniques are replaced by advanced architectures such as LSTM, RNN, and CNN, which demonstrate stronger representational capacity for complex financial signals. These models \cite{Wang2017Session,Xie2023,Wu2025,abd2025atad} deliver strong performance by accurately modeling temporal dependencies, sequential dynamics, and localized features within online transactional data. They also reveal subtle and irregular behavioral patterns that remain obscure to traditional approaches, which allows a more accurate assessment of hidden and evolving fraud risks.

However, traditional deep learning models still treat online transactions as independent samples and overlook cross-entity relationships. This limitation reduces their capability to capture coordinated or camouflaged fraud. As a result, researchers increasingly adopt GNNs, which model structural dependencies and multi-hop interaction patterns to reveal sophisticated web-based fraudulent behaviors more effectively.

\subsection{Graph-based Financial Risk Control}
Graph learning has proven highly effective \cite{rahman2024sacnn,wu2024safeguarding} in detecting web finance fraud by modeling the structural dependencies between entities \cite{Zhu2024}.
SA-GNN \cite{liu2023preventing} builds graphs via selective representation and dual similarity constraints based on features. Semi-GNN \cite{wang2019semi} employs a hierarchical attention mechanism to model diverse transaction behaviors. SplitGNN \cite{wu2023splitgnn} partitions the graph through edge classification and introduces adaptive filters to enhance representation for fraud-related anomalies. ASA-GNN \cite{tian2023asa} learns discriminative embeddings and samples informative neighbors to suppress noise. HHSGT \cite{wang2024homo} employs a relation-aware sparse graph transformer and addresses the structural heterogeneity observed in financial fraud graphs. MTPNet \cite{guang2025multi} builds transaction graphs at multiple temporal scales and applies adaptive neighborhood aggregation to highlight the most relevant behavioral cues, which ultimately leads to more accurate detection of fraudulent activities. FraudGCN \cite{wang2024multi} constructs a heterogeneous corporate network to integrate industrial, supply chain, and audit-sharing relationships. Moreover, it employs a graph encoder with relation-wise aggregation and attention fusion, supplemented by a diffusion-based sampling strategy and focal loss to handle class imbalance.

Nonetheless, these models provide limited support for uncovering imitation-based online fraudulent behaviors and lack dedicated mechanisms to effectively identify infrequent yet critical fraud instances in the tail of long-tailed distributions.

\section{Preliminaries}
In this section, we elaborate on two hippocampal mechanisms that inspire model design: the scene conflict detection mechanism and the match-mismatch novelty detection mechanism.
\subsection{Scene Conflict Detection Mechanism}
Scene conflict detection in hippocampus involves identifying inconsistencies across sensory inputs originating from distinct modalities \cite{julian2021remapping}. This enables hippocampus system to resolve spatial or contextual conflicts by integrating mismatched elements into a coherent scene representation.

For instance, when an individual hears a familiar voice but visually identifies an unfamiliar face, the conflict between auditory and visual modalities may trigger the hippocampal scene conflict detection. This process enables the brain to reconcile incongruent sensory inputs and identify potentially deceptive or incongruent environmental cues.
\subsection{Match-Mismatch Novelty Detection Mechanism}
In the hippocampal CA3 region, the brain forms predictions about sensory inputs based on memory representations. When actual inputs deviate from these predictions, CA1 region generates a mismatch response \cite{bittner2015conjunctive}, which serves as a neural marker of novelty, highlighting rare and unexpected patterns in the environment.

For example, when going home each day, CA3 predicts the familiar scene at the apartment entrance. If the sensory input matches this expectation, CA1 remains relatively inactive. However, if a cat unexpectedly appears on the shoe cabinet, CA1 generates a strong mismatch response, signaling novelty. Following such an event, one may begin to wonder whether a cat will appear at the apartment entrance each day and allocate heightened attention to a previously inconspicuous element, namely the shoe cabinet.

\section{Methodology}
In this section, we introduce the details of HIMVH. As shown in Figure 3, HIMVH consists of four modules: multi-view transaction hypergraph construction module, hippocampal cross-view discrepancy perception module, CA1-inspired novelty-aware hypergraph learning module, multi-view hypergraph fusion module.

\subsection{Multi-view Transaction Hypergraph Construction}
We first introduce the Multi-view Transaction Hypergraph Construction (MTHC) module. Specifically, we construct distinct views of the online transaction data based on different attribute columns such as sender ID, receiver ID, transaction location, and transaction type, each serving as a view-specific key, and the view set is defined as $\mathcal{A}$. For each view, we build undirected hypergraphs by applying a temporal sliding window over chronologically ordered transactions.

For a given view-specific key $key_a$, we define the corresponding view $a\in\mathcal{A}$ as an undirected hypergraph $\mathcal{G}^{(a)}=(\mathcal{V}^{(a)},\mathcal{E}^{(a)})$, where $\mathcal{V}^{(a)}$ denotes the set of nodes, with each node representing an individual transaction. The set of hyperedges $\mathcal{E}^{(a)}$ is constructed such that transactions falling within the same temporal sliding window form a hyperedge. In the following, we provide a detailed description of the hyperedge construction process in view $a$.
\begin{equation}
\forall c\in\mathcal{C}_a,\mathcal{V}^a_{c}=Sort_{time}(\{v_i\in\mathcal{V}|Cate^a(v_i)=c\}),
\label{eq 1}
\end{equation}
where $\mathcal{A}$ denotes the set of all views, and for each view $a\in\mathcal{A}$, $\mathcal{C}_a$ represents the set of unique categories under view $a$ (i.e., view-specific key $key_a$). $\mathcal{V}^a_{c}$ denotes the set of transactions belonging to category $c$, sorted in ascending order by their transaction timestamps.
\begin{equation}
e^a_{c,j}=\{v_{c_j},v_{c_{j+1}},\cdots,v_{c_{j+w-1}}\},j=1,2,\cdots,|\mathcal{V}^a_{c}|-w+1,
\label{eq 2}
\end{equation}
where $e^a_{c,j}$ denotes the $j$-th hyperedge constructed under view $a$ for category $c$, and $w$ is a hyperparameter that defines the size of the temporal sliding window, corresponding to the number of nodes contained within each hyperedge.

Therefore, the hyperedge set under view $a$ is defined as:
\begin{equation}
\mathcal{E}^{(a)}=\bigcup_{c\in\mathcal{C}_a}\bigcup_{j}e^a_{c,j}.
\label{eq 3}
\end{equation}

\subsection{Hippocampal Cross-view Discrepancy Perception}
As introduced in the Preliminaries, the hippocampus integrates cues from different sources and detects inconsistencies to identify web finance fraud. Inspired by its scene conflict detection mechanism, we propose a Hippocampal Cross-view Discrepancy Perception (HCDP) module to detect camouflaged fraud. Although fraudsters may mimic normal transaction behavior, certain attributes such as transaction location or IP address are difficult to manipulate. This leads to inconsistencies across views, which the HCDP module captures to reveal subtle signs of deception.

We design multi-dimensional metrics to quantify the inconsistencies of the same transaction node across different views, including identity ($ID\_Diff$), feature ($Feat\_Diff$), and label discrepancy ($Label\_Diff$).
\begin{equation}
ID\_Diff^{(a_1,a_2)}_i=1-\frac{|\mathcal{N}^{a_1}_i\bigcap\mathcal{N}^{a_2}_i|}{|\mathcal{N}^{a_1}_i\bigcup\mathcal{N}^{a_2}_i|},
\label{eq 4}
\end{equation}
where $\mathcal{N}^{a_1}_i$ and $\mathcal{N}^{a_2}_i$ denote the sets of neighboring nodes of node $i$ in views $a_1$ and $a_2$, respectively. The term $ID\_Diff^{(a_1,a_2)}_i$ measures the structural discrepancy of node $i$ across views $a_1$ and $a_2$, and is computed as the Jaccard distance between its two neighbor sets.
\begin{equation}
p^a_i=Softmax(Mean(\{x_j|j\in\mathcal{N}_i^a\})),
\label{eq 5}
\end{equation}
\begin{equation}
Feat\_Diff^{(a_1,a_2)}_i=\frac{1}{2}\cdot KL(M||p_i^{a_1})+\frac{1}{2}\cdot KL(M||p_i^{a_2}),
\label{eq 6}
\end{equation}
where $x_j$ is the feature of neighbor $j$ under view $a$, and $p_i^a$ represents the distribution of neighboring features. In Eq. (6), $Feat\_Diff^{(a_1,a_2)}$ denotes the feature discrepancy between views $a_1$ and $a_2$ with Jensen–Shannon divergence, and $M$ is the average of the two distributions $p_i^{a_1}$ and $p_i^{a_2}$.
\begin{equation}
\mathcal{H}^{a}_i=-\sum_{label\in\{0,1\}}r^a_{i,label}\cdot\log(r^a_{i,label}+\epsilon),
\label{eq 7}
\end{equation}
\begin{equation}
Label\_Diff^{(a_1,a_2)}_i=|\mathcal{H}^{a_1}_i-\mathcal{H}^{a_2}_i|,
\label{eq 8}
\end{equation}
where $\mathcal{H}^{a}_i$ denotes the label entropy of node $i$ under view $a$, and $r^a_{i,label}$ is the empirical probability of each label (0 for normal, 1 for fraudulent) among its neighbors. $Label\_Diff^{(a_1,a_2)}_i$ reflects the inconsistency of local label distributions between views $a_1$ and $a_2$. To address potential label leakage during inference, we compute label entropy using the predicted risk distribution from a lightweight auxiliary MLP instead of ground-truth neighbor labels.
\begin{multline}
h_i = \text{Concat}\Bigl[x_i,
\frac{1}{\binom{|\mathcal{A}|}{2}}\sum_{j<k}\text{ID\_Diff}^{(a_j,a_k)}_i,\\
\frac{1}{\binom{|\mathcal{A}|}{2}}\sum_{j<k}\text{Feat\_Diff}^{(a_j,a_k)}_i,
\frac{1}{\binom{|\mathcal{A}|}{2}}\sum_{j<k}\text{Label\_Diff}^{(a_j,a_k)}_i\Bigr],
\label{eq 9}
\end{multline}
where $h_i$ denotes the enhanced representation of node $i$, formed by concatenating its original feature $x_i$ with three types of cross-view discrepancy metrics: identity, feature, and label discrepancy. Each discrepancy term is computed by averaging over all $\binom{|\mathcal{A}|}{2}$ unordered pairs of views.

\subsection{CA1-inspired Novelty-aware Hypergraph Learning}
Inspired by the match-mismatch novelty detection in the Preliminaries, we propose the CA1-inspired Novelty-aware Hypergraph Learning (CNHL) module, a heterophily-sensitive graph learning paradigm. It adaptively reweights messages based on local deviation to enhance the representation of long-tailed samples. We illustrate it by detailing a single layer of CNHL.

\begin{equation}
\mu_e=\frac{1}{w}\sum_{i\in e}h_i,
\label{eq 10}
\end{equation}
where $w$ denotes both the temporal window size and the number of nodes within hyperedge $e$, and $\mu_e$ represents the hyperedge center of all $w$ nodes. The center serves as an analogy to the predictive pattern generated by CA3 region. It is noted that HIMVH incorporates label propagation for semi-supervised scenarios, with label embeddings for labeled nodes and zero embeddings for unlabeled ones. The node representation $h_i$ is formed by fusing the enhanced features with the label embeddings, followed by the normalization process.
\begin{equation}
Var_l=\frac{1}{k}\sum_{i\in e}(h_{i,l}-\mu_{e,l})^2,
\label{eq 11}
\end{equation}
\begin{equation}
g_l=\frac{exp(\beta\cdot Var_l)}{\sum_{m=1}^d exp(\beta\cdot Var_m)},
\label{eq 12}
\end{equation}
where the variance term $Var_l$ measures novelty in the $l$-th feature dimension. Similar to CA1’s strong mismatch response to unexpected changes (e.g., a cat on the shoe cabinet), higher variance signals greater deviation from expected patterns within the hyperedge. Such dimensions are deemed more novel and receive larger attention weights $g_l$, enabling the model to focus on informative atypical features. The parameter $\beta$ controls sensitivity to these deviations.
\begin{equation}
s_i=\sum_{l=1}^d g_l\cdot(h_{i,l}-\mu_{e,l})^2,
\label{eq 13}
\end{equation}
\begin{equation}
m_i=\sum_{j\in e\setminus{i}}\alpha_j\cdot Wh_j,
\label{eq 14}
\end{equation}
where $s_i$ represents the novelty score of node $i$, and $\alpha_j$ is the normalized novelty score derived from $s_j$; $h_{i,l}$ is the $l$-th feature dimension of node $i$, and $\mu_{e,l}$ denotes the $l$-th dimension of the hyperedge center. In addition, the term $m_i$ denotes the aggregated message passed to node $i$.

The update process of CNHL can be defined as follows:
\begin{equation}
h^\prime_i=Sigmoid(h_i+m_i).
\label{eq 15}
\end{equation}


\subsection{Multi-View Hypergraph Fusion}
In this section, we perform weighted fusion of transaction node embeddings from different views for Multi-view Hypergraph Fusion (MHF), followed by a downstream fraud detection task. This allows the model to integrate complementary information across heterogeneous perspectives.

\begin{equation}
\alpha_{vs}=Softmax(W_g\cdot Relu(Concat[h^\prime_{a_1},h^\prime_{a_2},\cdots,h^\prime_{a_{|\mathcal{A}|}}])),
\label{eq 16}
\end{equation}
where $h^\prime_{a_1}, h^\prime_{a_2}, \ldots, h^\prime_{a_{|\mathcal{A}|}}$ denote the learned embeddings of a transaction node under each view, and $\alpha_{vs}$ represents the attention weights across views, used to adaptively fuse multi-view representations.
\begin{equation}
\delta\alpha_{vs}^{(a)}=MLP^{(a)}(Tanh(W_p\cdot\sum_{j\in\mathcal{A}}h^{\prime}_{j})),
\label{eq 17}
\end{equation}
where $\delta\alpha_{vs}^{(a)}$ represents the view-specific modulation term, used to refine the final fusion weights.
\begin{equation}
h^{final}=\sum_{a=1}^{\mathcal{\mathcal{|A|}}}\frac{|\alpha^{(a)}_{vs}+\delta\alpha_{vs}^{(a)}|}{\sum_{j=1}^{|\mathcal{A}|}(|\alpha^{(j)}_{vs}+\delta\alpha_{vs}^{(j)}|)}\cdot h^{\prime}_{a},
\label{eq 18}
\end{equation}
where the final fused representation $h^{final}$ is obtained via a reweighted combination across views, with modulation-enhanced attention normalized across all views. Here, $\alpha^{(a)}_{vs} + \delta\alpha^{(a)}_{vs}$ denotes the adjusted view weight after modulation, which is normalized to derive the final view-level attention.

The representation $h^{final}$ serves as the input to the final classifier. The classifier is a lightweight MLP, which includes a linear transformation, followed by batch normalization, activation, dropout, and another linear transformation. HIMVH is trained end-to-end, with the classification cross-entropy loss guiding the optimization of all components.

The process of HCDP and CNHL is presented in Algorithm 1.
\begin{algorithm}[H]
\caption{Brain-inspired fraud detection mechanisms}
\label{alg:CDP and NAH}

\begin{algorithmic}[1]
\Input Constructed multi-view hypergraph $\mathcal{G}$.
\Output Embedding result $h^\prime_i$ for node $i$ in each view.
\Statex // Cross-view discrepancy perception
\algnotext{ // Cross-view discrepancy perception}
\For{each node $i \in \mathcal{V}$}:
    \For{each unordered view pair $(a_j, a_k)$, $j < k$}:
        \State $ID\_Diff^{(a_1,a_2)}_i=1-\frac{|\mathcal{N}^{a_1}_i\bigcap\mathcal{N}^{a_2}_i|}{|\mathcal{N}^{a_1}_i\bigcup\mathcal{N}^{a_2}_i|}$;
        \State Compute $Feat\_Diff^{(a_1,a_2)}_i$ by Eqs. (5) and (6); 
        \State Obtain $Label\_Diff^{(a_1,a_2)}_i$ by Eqs. (7) and (8);
    \EndFor
    \State Obtain the enhanced $h_i$ by concatenating the three 
    \Statex \hspace*{1.4em} differences with original features $x_i$ by Eq. (9);
\EndFor
\Statex // Novelty-aware hypergraph neural network
\For{\(n \gets 1\) to $N$}: // The layer of hypergraph
    \For{each node $i \in \mathcal{V}$}:
        \State Compute the hyperedge center $\mu_e$;
        \State Compute the novelty weight $g_l$;
        \State $s_i=\sum_{l=1}^d g_l\cdot(h_{i,l}-\mu_{e,l})^2$; // Novelty score
        \State Obtain the aggregated message by Eq. (14);
        \State Update node embedding by Eq. (15);
    \EndFor
\EndFor
\end{algorithmic}
\end{algorithm}

\section{Experiments}
\subsection{Experimental Setup}
\textbf{Datasets.}\quad We conduct experiments on six financial fraud datasets that reflect real risks faced by online users, including two public and four private datasets. These datasets represent fraud patterns that threaten societal fairness and the security of web-based financial ecosystems. The statistical details are summarized in Table 1.

\begin{table}[h]
\renewcommand\arraystretch{1.0}
\centering
\caption{Statistics of six datasets.}
\begin{tabular}{cccccccccccccc}
\toprule[1pt] 
\multicolumn{3}{c}{Dataset} & \multicolumn{3}{c}{\#Record}  & \multicolumn{2}{c}{\#Fraud} &  \multicolumn{3}{c}{\#Legitimate} & \multicolumn{3}{c}{\#Unlabeled}   \\

\toprule[1pt] 

\multicolumn{3}{c}{S-FFSD} & \multicolumn{3}{c}{77,881} & \multicolumn{2}{c}{5,256} & \multicolumn{3}{c}{24,387} & \multicolumn{3}{c}{48,238} \\

\multicolumn{3}{c}{Sparkov} & \multicolumn{3}{c}{99,728} & \multicolumn{2}{c}{298} & \multicolumn{3}{c}{29,620} & \multicolumn{3}{c}{69,810} \\

\multicolumn{3}{c}{Private-1} & \multicolumn{3}{c}{666,592} & \multicolumn{2}{c}{10,478} & \multicolumn{3}{c}{56,181} & \multicolumn{3}{c}{599,933} \\

\multicolumn{3}{c}{Private-2} & \multicolumn{3}{c}{691,661} & \multicolumn{2}{c}{609} & \multicolumn{3}{c}{33,974} & \multicolumn{3}{c}{657,078} \\

\multicolumn{3}{c}{Private-3} & \multicolumn{3}{c}{264,807} & \multicolumn{2}{c}{313} & \multicolumn{3}{c}{39,408} & \multicolumn{3}{c}{225,086} \\

\multicolumn{3}{c}{Private-4} & \multicolumn{3}{c}{1,243,035} & \multicolumn{2}{c}{324} & \multicolumn{3}{c}{49,397} & \multicolumn{3}{c}{1,193,314} \\

\bottomrule[1pt]
\end{tabular}

\end{table}

\begin{table*}[htbp]
\renewcommand\arraystretch{0.933}
\centering
\label{tab:results}
\small
\caption{Fraud detection performance (\%) on six datasets compared with 15 SOTA models.}
\setlength{\tabcolsep}{6pt}
\begin{tabular}{c*{7}{c@{\hspace{4pt}}c@{\hspace{4pt}}c}}
\toprule

\multirow{2}{*}{Method} & \multicolumn{3}{c}{S-FFSD} & \multicolumn{3}{c}{Sparkov} & \multicolumn{3}{c}{Private-1} & \multicolumn{3}{c}{Private-2} & \multicolumn{3}{c}{Private-3} & \multicolumn{3}{c}{Private-4} \\
\cmidrule(r){2-4} \cmidrule(lr){5-7} \cmidrule(lr){8-10} \cmidrule(lr){11-13} \cmidrule(lr){14-16} \cmidrule(lr){17-19} \cmidrule(l){20-22}
 & AUC & F1 & AP & AUC & F1 & AP & AUC & F1 & AP & AUC & F1 & AP & AUC & F1 & AP & AUC & F1 & AP \\
\midrule
Decision Tree & 67.71 & 72.76 & 45.51 & 67.12 & 74.54 & 31.24 & 85.69 & 85.27 & 60.28 & 89.24 & 87.32 & 56.99 & 85.75 & 84.51 & 48.25 & 76.42 & 79.04 & 34.68\\
XGBoost & 69.61 & 74.88 & 48.73 & 67.82 & 75.05 & 31.54 & 86.54 & 86.91 & 64.21 & 76.31 & 81.85 & 44.16 & 81.23 & 83.21 & 44.92 & 74.79 & 79.56 & 36.92\\
\midrule
MCNN & 75.50 & 64.81 & 32.85 & 86.17 & 65.26 & 15.62 & 93.50 & 88.77 & 68.70 & 96.31 & 88.80 & 62.78 & 93.19 & 81.19 & 42.86 & 96.49 & 79.80 & 41.40 \\
STAN & 76.82 & 72.09 & 38.52 & 88.90 & 63.79 & 14.91 & 93.36 & 87.16 & 65.27 & 96.22 & 84.05 & 51.16 & 93.25 & 81.39 & 43.30 & 93.07 & 78.25 & 36.87 \\
\midrule
PNA & 69.73 & 63.48 & 29.11 & 86.04 & 78.59 & 35.09 & 91.26 & 81.14 & 48.52 & 96.63 & 84.22 & 51.77 & 88.93 & 83.09 & 45.75 & 85.99 & 78.32 & 34.47 \\
STAGN & 80.03 & 74.01 & 41.72 & 94.58 & 83.55 & 51.03 & 94.62 & 89.83 & 71.37 & 98.64 & 82.20 & 48.39 & 92.21 & 79.41 & 38.93 & 93.65 & 77.81 & 36.29 \\
PC-GNN & 85.44 & 70.31 & 69.37 & 92.53 & 55.05 & 39.57 & 97.75 & 86.17 & 85.07 & 98.24 & 67.61 & 71.39 & 95.76 & 68.72 & 43.77 & 98.60 & 57.23 & 58.69 \\
BWGNN & 87.28 & 72.78 & 71.66 & 96.59 & 83.98 & 78.66 & 98.18 & 91.00 & 86.57 & 98.52 & 85.76 & 79.14 & 97.49 & 73.60 & 53.47 & 98.62 & 74.81 & 56.31 \\
GHRN & 87.44 & 72.04 & 72.12 & 97.52 & 84.99 & 78.82 & 98.17 & 91.27 & 86.42 & 98.79 & 87.63 & 79.57 & 96.21 & 78.13 & 52.70 & 96.51 & 79.14 & 56.75 \\
GTAN & 82.86 & 73.36 & 65.85 & 96.24 & 77.74 & 64.22 & 98.29 & 92.89 & 88.19 & 98.29 & 87.93 & 79.89 & 98.67 & 85.60 & 75.56 & 96.56 & 79.08 & 64.35 \\
U-A2GAD & 85.70 & 76.82 & 70.07 & 94.94 & 82.49 & 67.36 & 97.69 & 91.49 & 85.32 & 97.44 & 87.10 & 69.29 & 98.68 & 80.23 & 51.76 & 97.48 & 76.96 & 45.01 \\
ConsisGAD & 72.18 & 71.73 & 55.35 & 88.01 & 68.91 & 35.54 & 97.83 & 91.44 & 85.88 & 97.69 & 87.21 & 64.56 & 97.23 & 78.59 & 38.57 & 96.38 & 76.72 & 36.92 \\
UniGAD & 84.92 & 76.29 & 69.18 & 96.89 & 81.57 & 62.23 & 98.25 & 92.65 & 87.57 & 99.00 & 81.71 & 78.61 & 98.49 & 81.58 & 67.26 & 98.05 & 78.97 & 52.78 \\
RGTAN & 84.61 & 75.13 & 69.39 & 95.81 & 81.43 & 67.22 & 98.37 & 92.51 & 89.43 & 98.63 & 86.35 & 79.52 & 98.16 & 83.50 & 71.39 & 97.44 & 78.88 & 64.16 \\
SpaceGNN & 86.55 & 75.89 & 70.27 & 97.62 & 85.27 & 79.99 & 98.09 & 92.07 & 87.24 & 98.93 & 87.63 & 81.60 & 98.45 & 85.34 & 75.99 & 97.14 & 77.59 & 56.42 \\
HIMVH & \makebox[0pt][c]{\textbf{88.83}} & \makebox[0pt][c]{\textbf{78.74}} & \makebox[0pt][c]{\textbf{73.56}} & \makebox[0pt][c]{\textbf{98.21}} & \makebox[0pt][c]{\textbf{93.79}} & \makebox[0pt][c]{\textbf{87.76}} & \makebox[0pt][c]{\textbf{98.64}} & \makebox[0pt][c]{\textbf{92.91}} & \makebox[0pt][c]{\textbf{90.61}} & \makebox[0pt][c]{\textbf{99.17}} & \makebox[0pt][c]{\textbf{90.45}} & \makebox[0pt][c]{\textbf{85.68}} & \makebox[0pt][c]{\textbf{99.36}} & \makebox[0pt][c]{\textbf{86.34}} & \makebox[0pt][c]{\textbf{78.82}} & \makebox[0pt][c]{\textbf{99.00}} & \makebox[0pt][c]{\textbf{84.32}} & \makebox[0pt][c]{\textbf{72.28}}\\
\bottomrule
\end{tabular}
\label{table2}
\end{table*}

Xiang et al. \cite{xiang2023semi} collect online records from collaborative partners and name it as S-FFSD\footnote{https://github.com/AI4Risk/antifraud}. The Sparkov\footnote{https://github.com/namebrandon/Sparkov} dataset is publicly available, with a record of transactions of 100 customers from February 1, 2023, to March 31, 2023. We employ a dataset of web finance transaction records provided by a financial institution as Private, spanning from January to April. The subsets labeled as Private-1 to Private-4 correspond to the online transaction data of each month. In addition to identity-related attributes of the transaction parties, each private dataset includes eight attribute features, such as indicators of IP address legitimacy, transaction timestamps, and other behavioral information.\\
\textbf{Baselines.}\quad We employ 15 SOTA models, which are mainly divided into three categories. Decision Tree \cite{xu2023efficient} and XGBoost \cite{nti2022scalable} are traditional models, MCNN \cite{fu2016credit} and STAN \cite{cheng2020spatio} are deep learning models without graph learning, while PNA \cite{xu2018powerful}, STAGN \cite{cheng2020graph}, PC-GNN \cite{liu2021pick}, BWGNN \cite{tang2022rethinking}, GHRN \cite{gao2023addressing}, GTAN \cite{xiang2023semi}, U-A2GAD \cite{li2025universal}, ConsisGAD \cite{chen2024consistency}, UniGAD \cite{lin2024unigad}, RGTAN \cite{xiang2025enhancing}, and SpaceGNN \cite{dong2025spacegnn} are graph learning models.\\
\textbf{Evaluation Metrics and Implementation Details.}\quad We utilize the area under curve (AUC), macro average of F1 score (F1), average precision (AP) to evaluate the effectiveness of our model. We implement HIMVH using PyTorch 1.13.1 and conduct all experiments on a server equipped with an NVIDIA V100 (32GB), Intel Xeon Gold 6230 CPUs. We employ the Adam optimizer with a learning rate of 0.001 and train the model for 50 epochs. We set batch size and hidden dimension to 512 and 256, respectively. We split the datasets chronologically into training, validation, and test sets with ratios of 60\%, 10\%, and 30\%. In particular, the number of views is 4, and the number of GNN layers is 3. Furthermore, the size of the temporal sliding window $w$ is set to 4, and novelty-sensitivity parameter $\beta$ is set to 1.

\subsection{Overall Results}
We comprehensively evaluate the detection performance of different models, and the results are shown in Table 2.

HIMVH consistently outperforms all baselines across six datasets in terms of three metrics. Compared with other graph learning baselines, HIMVH achieves average gains of 3.32\% in AUC, 9.36\% in F1 score, and 27.80\% in AP, confirming its superior generalization capability and effectiveness. These improvements stem from two components. HCDP detects latent inconsistencies in fraud across multiple views, effectively addressing the heterogeneity of camouflaged patterns. Meanwhile, CNHL captures deviations from local neighborhood predictions, enhancing the representation of sparse minority instances under tail regions.

Even on highly imbalanced datasets such as Private-3 and Private-4, HIMVH outperforms all baselines by at least 3.72\% and 12.32\% in AP, respectively. This advantage is largely attributed to the heterophily-aware graph learning in HIMVH, which enables more accurate message aggregation across dissimilar nodes and enhances the detection of underrepresented fraudulent patterns.

\subsection{Ablation Study}
\begin{figure}[H]
  \centering
    \begin{subfigure}{0.21\textwidth}
        \centering
        \includegraphics[width=\linewidth]{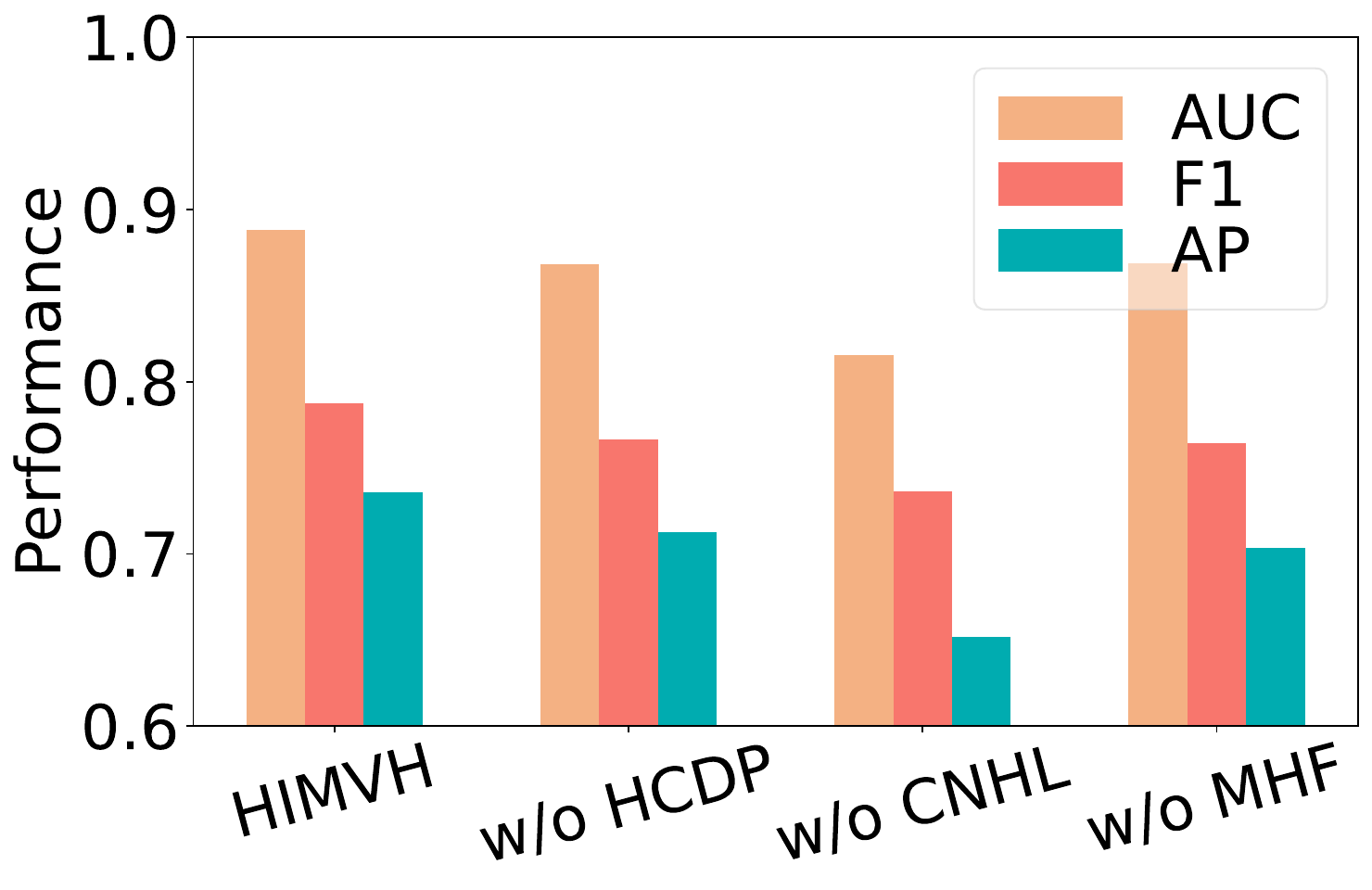}
        \caption{S-FFSD}
        \label{fig}
    \end{subfigure}
    \begin{subfigure}{0.21\textwidth}
        \centering
        \includegraphics[width=\linewidth]{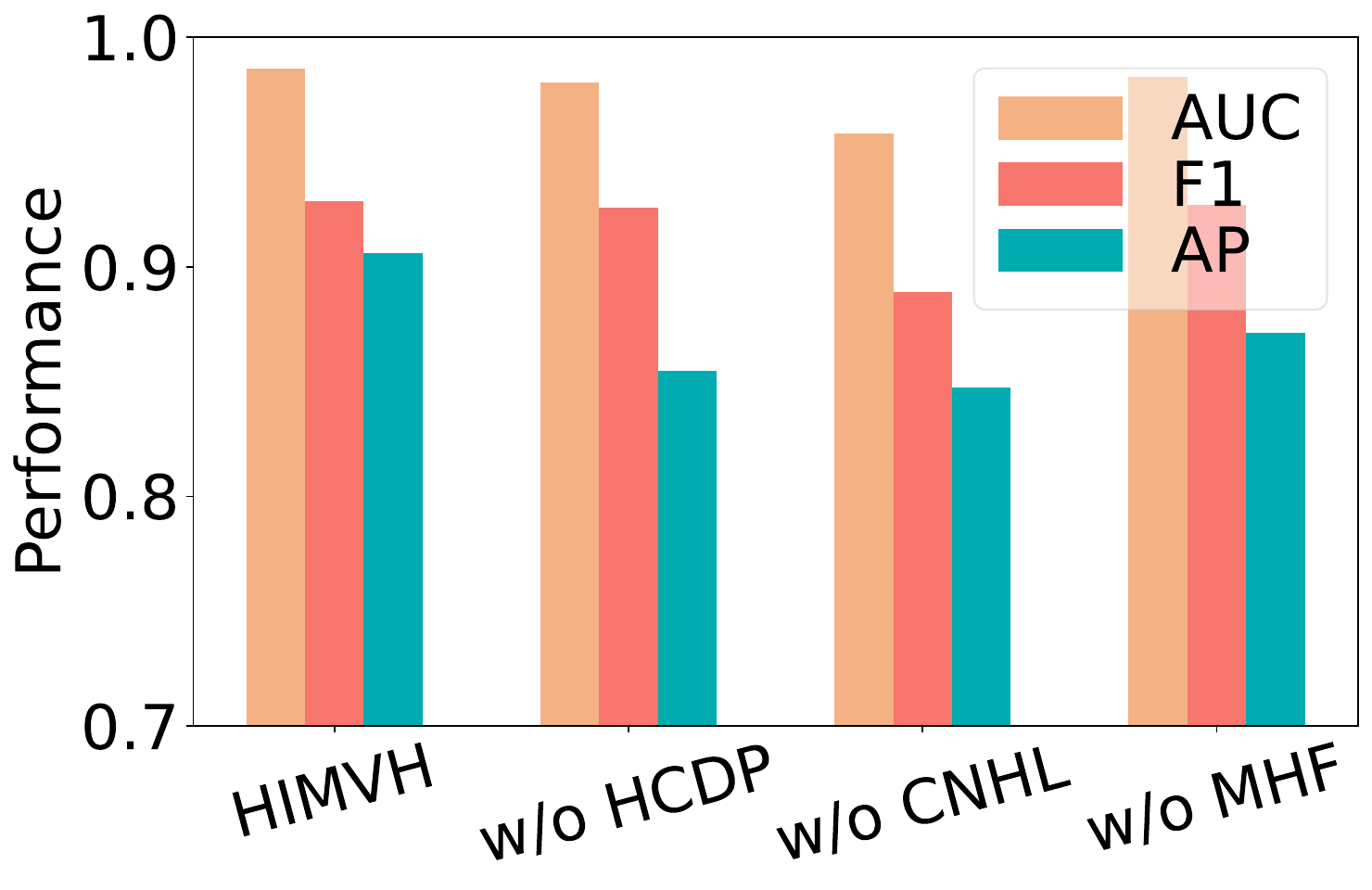}
        \caption{Private-1}
        \label{fig}
    \end{subfigure}
  \caption{The ablation analysis on S-FFSD and Private-1.}
  \label{Figure3}
\end{figure}

To examine the contribution of key design in HIMVH, we conduct ablation studies on three variants:
\begin{itemize}
    \item \textbf{w/o HCDP}: We remove the Hippocampal Cross-view Discrepancy Perception module.
    \item \textbf{w/o CNHL}: The component of CA1-inspired Novelty-aware Hypergraph Learning is replaced with multi-head attention mechanism.
    \item \textbf{w/o MHF}: We remove the Multi-view Hypergraph Fusion module and report the average classification performance across individual views as the final result.
\end{itemize}

As shown in Figure 4, HIMVH outperforms all three ablated variants, demonstrating the effectiveness of its core components. Specifically, the performance drops observed in w/o HCDP and w/o MHF confirm the importance of multi-view hypergraph modeling, particularly the role of cross-view inconsistency detection in uncovering camouflaged online fraudulent behaviors. Moreover, HIMVH w/o CNHL yields the poorest performance, as the heterophily-aware graph learning mechanism in HIMVH enables more adaptive and discriminative message propagation across structurally diverse nodes, which is critical for capturing minority fraud patterns under long-tailed distributions.

\begin{figure}[t]
  \centering
    \begin{subfigure}{0.21\textwidth}
        \centering
        \includegraphics[width=\linewidth]{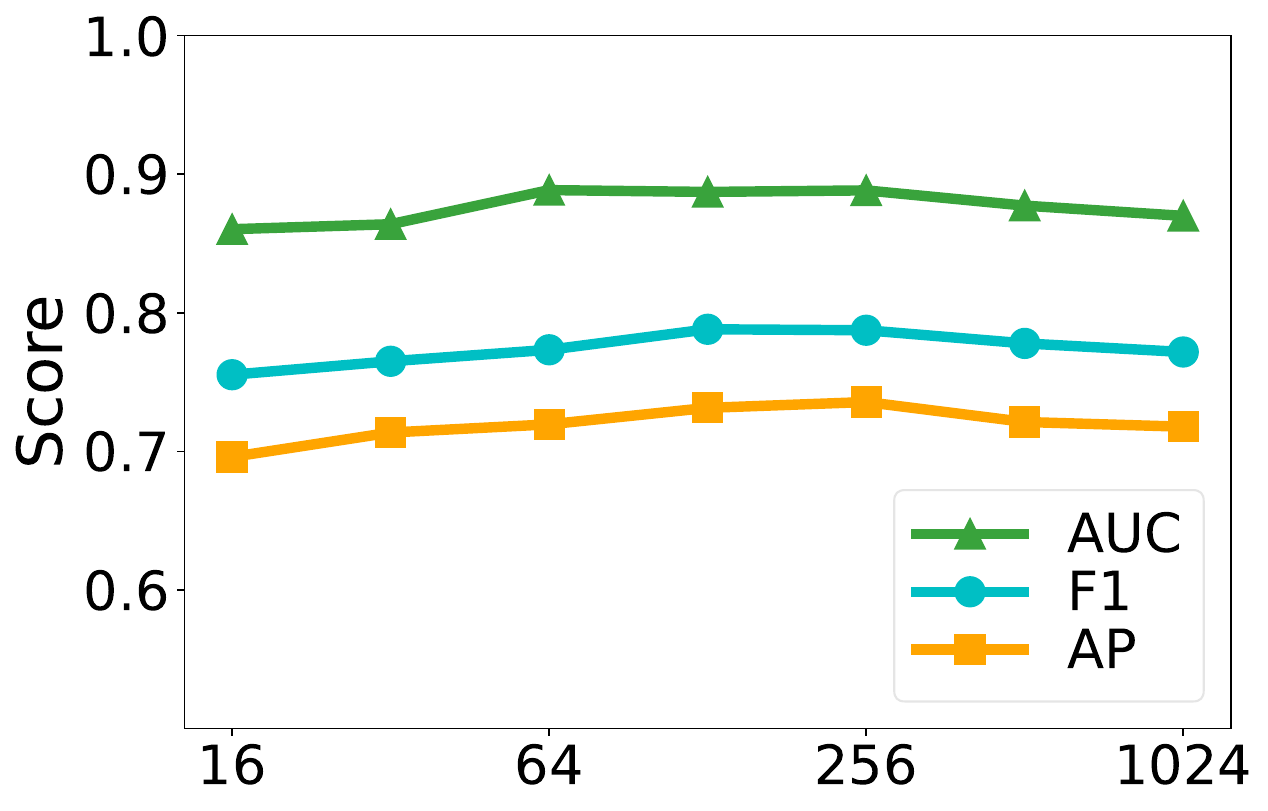}
        \caption{Hidden dimension}
        \label{fig}
    \end{subfigure}
    \begin{subfigure}{0.21\textwidth}
        \centering
        \includegraphics[width=\linewidth]{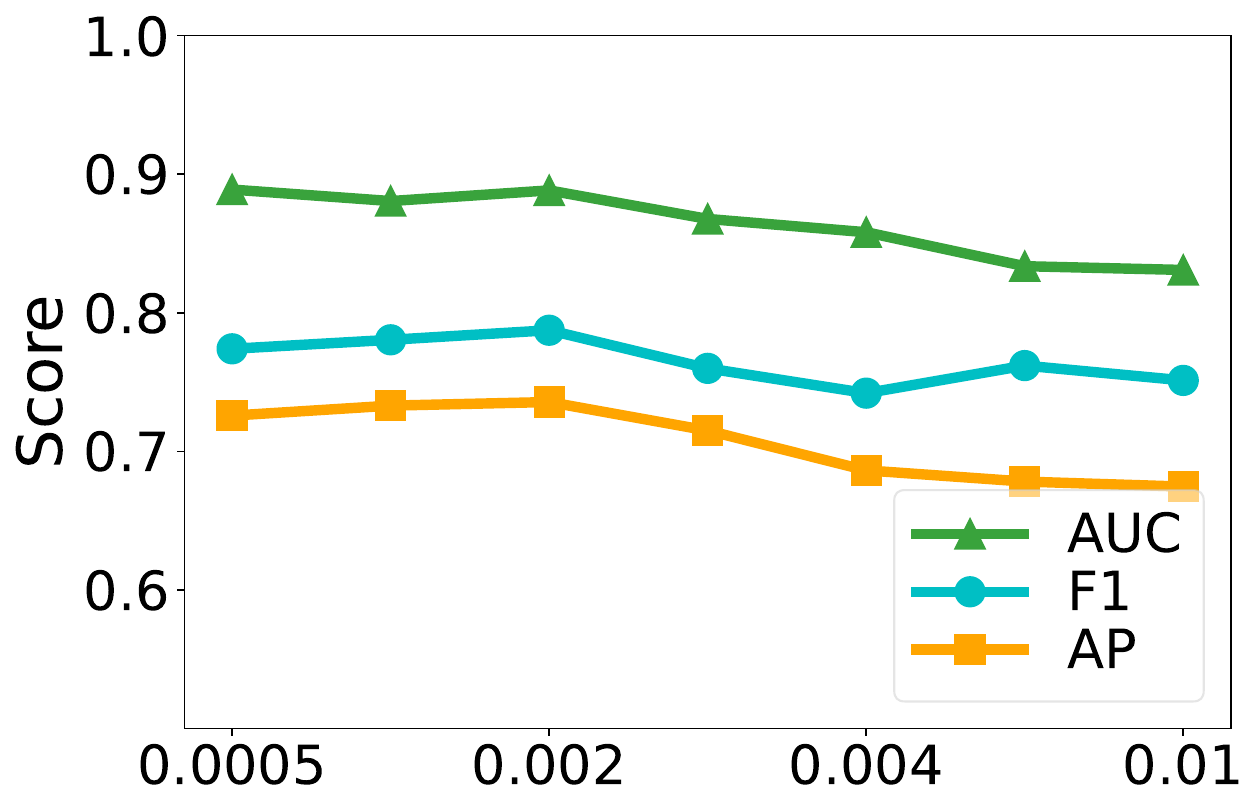}
        \caption{Learning rate}
        \label{fig}
    \end{subfigure}
    \begin{subfigure}{0.21\textwidth}
        \centering
        \includegraphics[width=\linewidth]{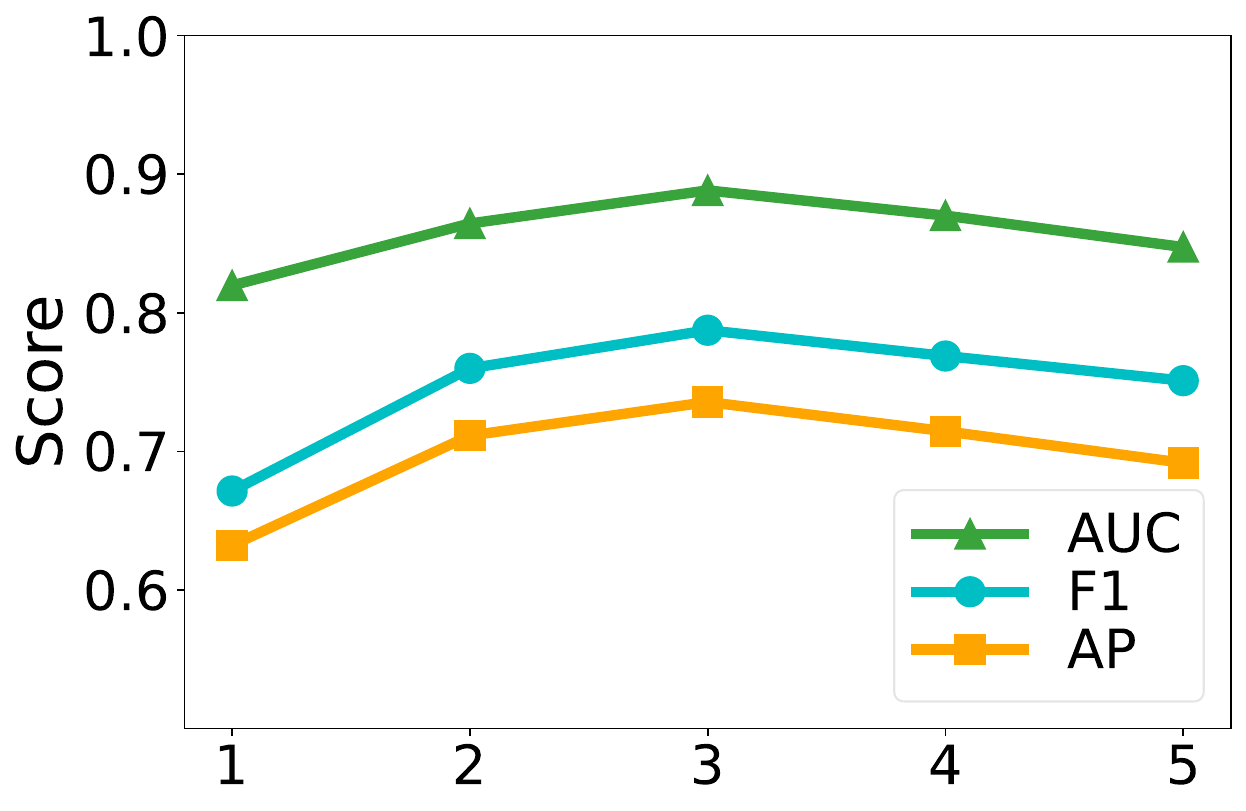}
        \caption{GNN layer}
        \label{fig}
    \end{subfigure}
    \begin{subfigure}{0.21\textwidth}
        \centering
        \includegraphics[width=\linewidth]{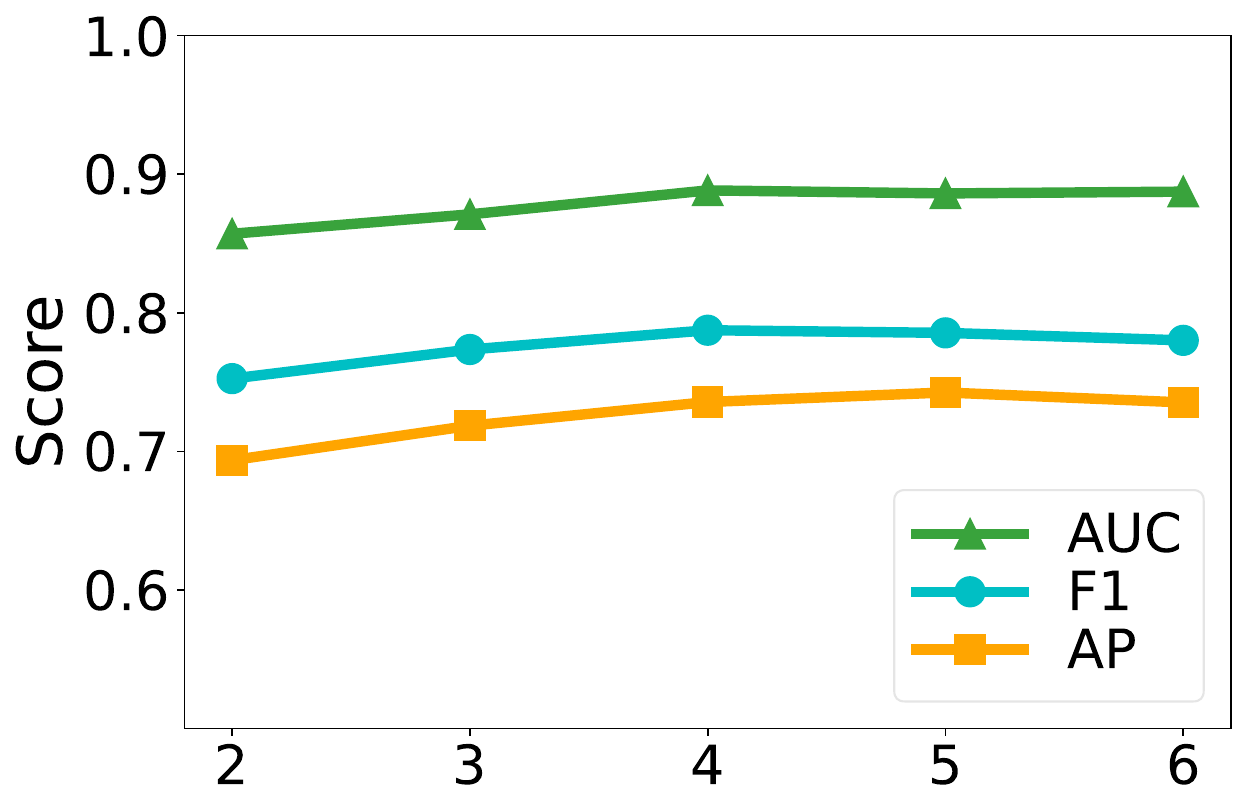}
        \caption{Window size}
        \label{fig}
    \end{subfigure}
  \caption{Parameter sensitivity analysis.}
  \label{Figure3}
\end{figure}
\subsection{Sensitivity Analysis}
To assess the robustness of HIMVH, we conduct a comprehensive sensitivity analysis on four key hyperparameters.

\begin{figure*}[htbp] 
  \centering
    \begin{subfigure}{0.24\textwidth}
        \centering
        \includegraphics[width=\linewidth]{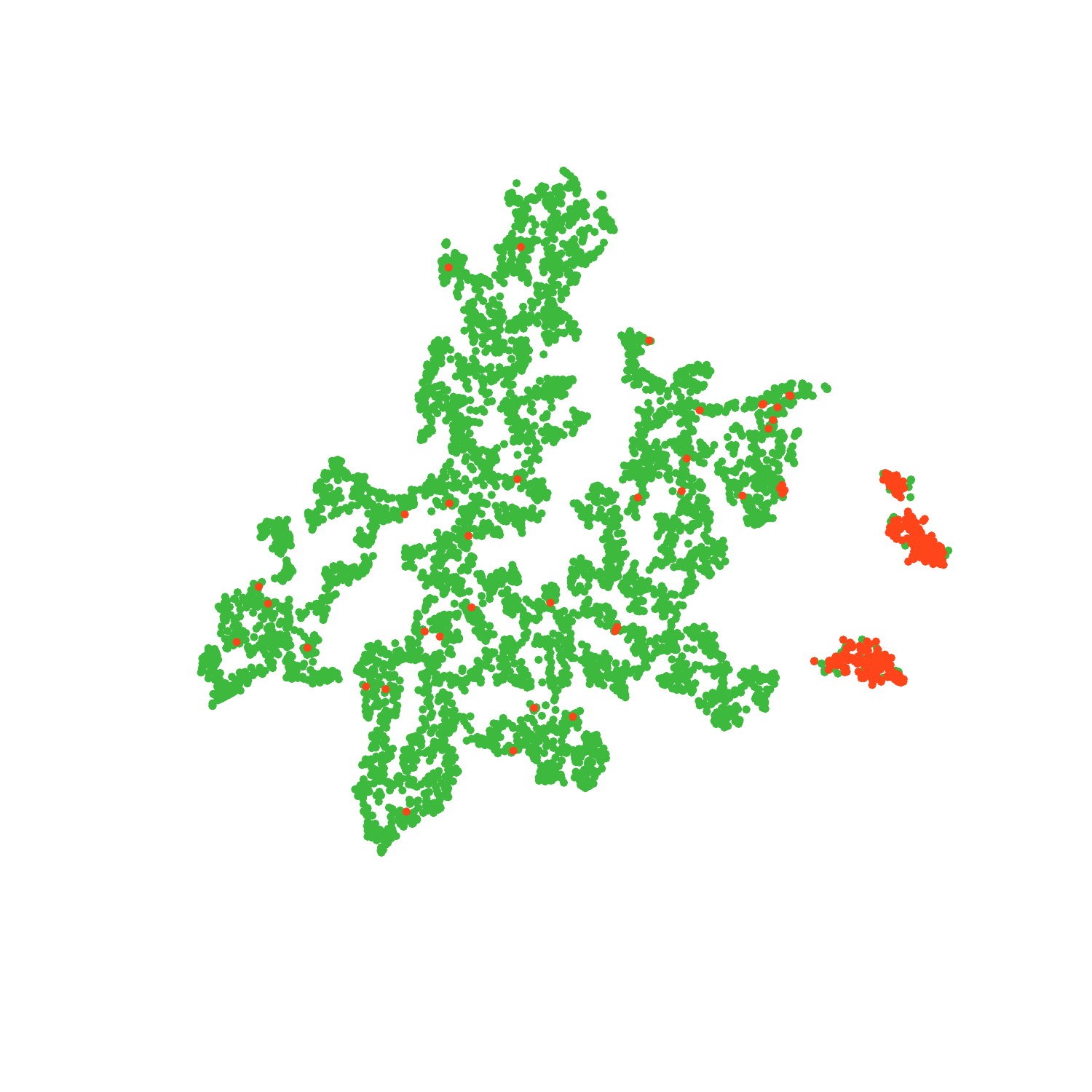}
        \caption{BWGNN}
        \label{fig}
    \end{subfigure}
    \begin{subfigure}{0.24\textwidth}
        \centering
        \includegraphics[width=\linewidth]{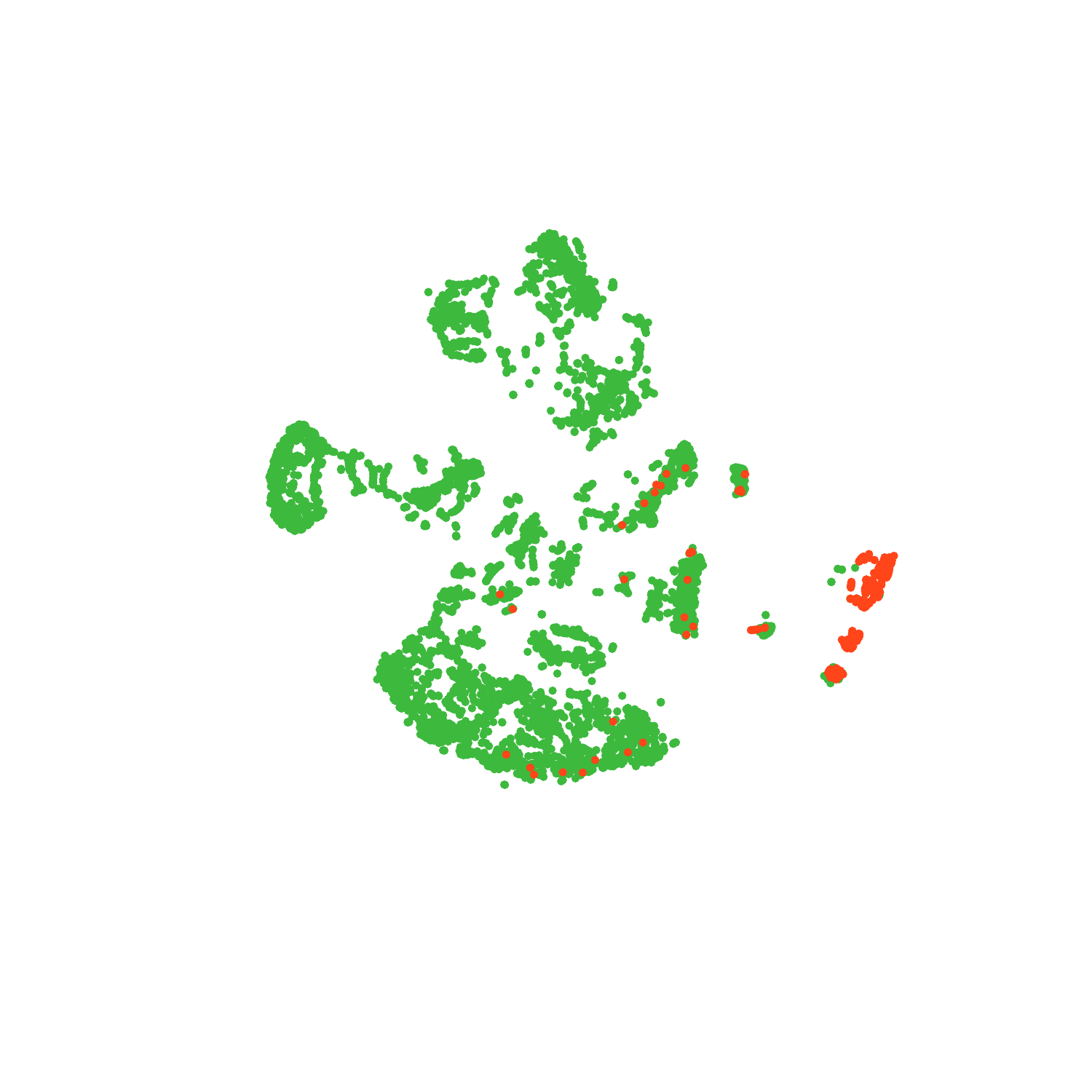}
        \caption{PCGNN}
        \label{fig}
    \end{subfigure}
    \begin{subfigure}{0.24\textwidth}
        \centering
        \includegraphics[width=\linewidth]{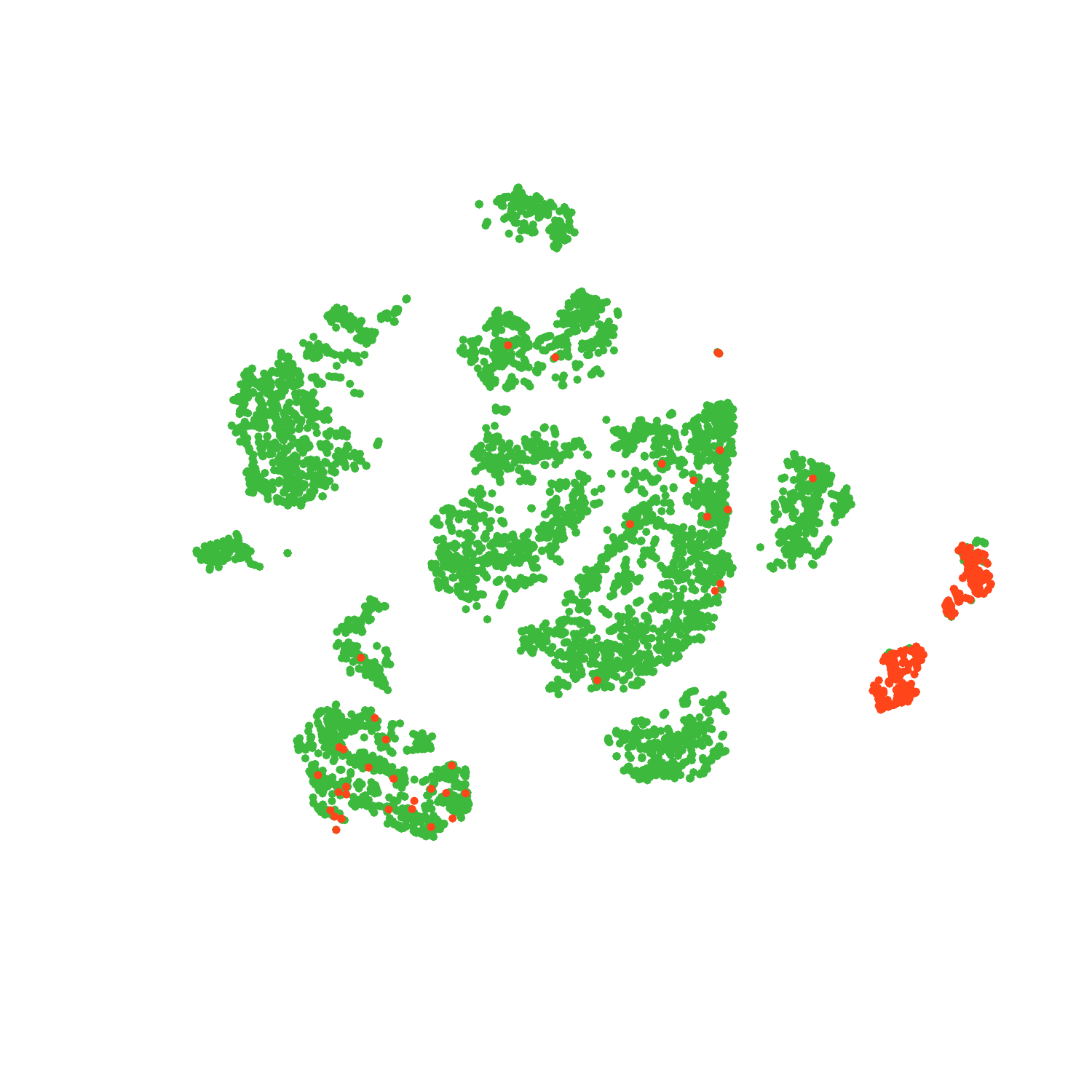}
        \caption{GHRN}
        \label{fig}
    \end{subfigure}
    \begin{subfigure}{0.24\textwidth}
        \centering
        \includegraphics[width=\linewidth]{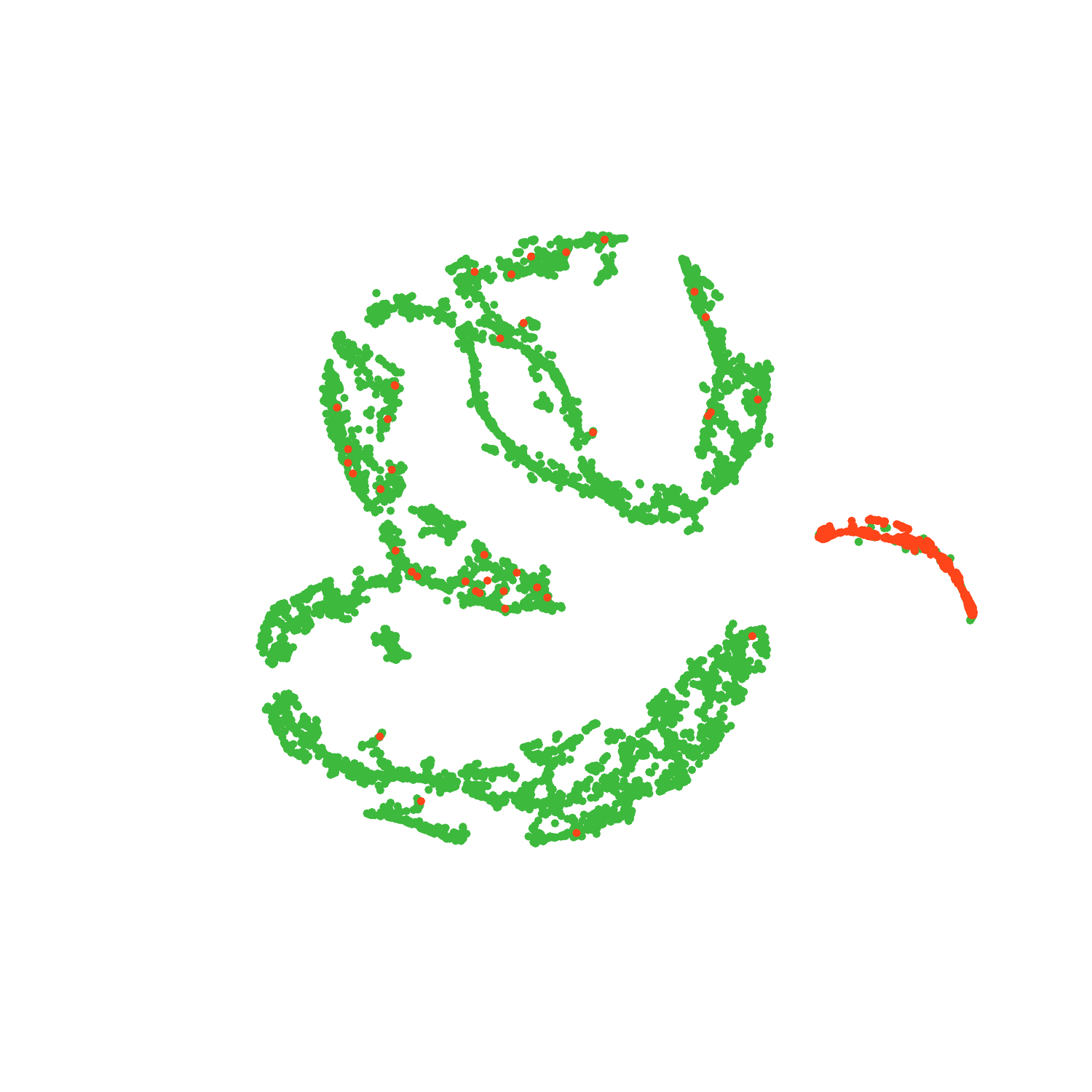}
        \caption{GTAN}
        \label{fig}
    \end{subfigure}
    \begin{subfigure}{0.24\textwidth}
        \centering
        \includegraphics[width=\linewidth]{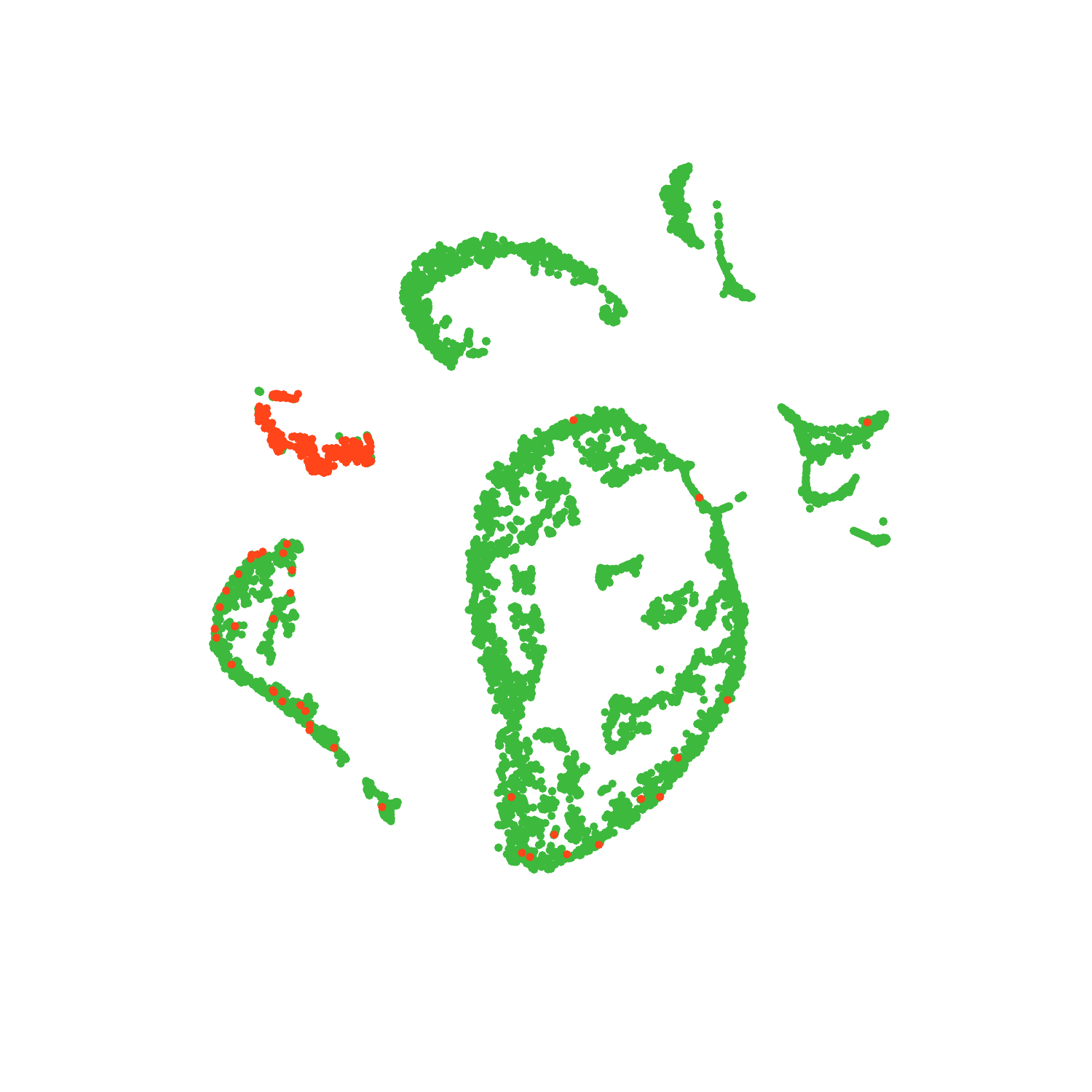}
        \caption{U-A2GAD}
        \label{fig}
    \end{subfigure}
    \begin{subfigure}{0.24\textwidth}
        \centering
        \includegraphics[width=\linewidth]{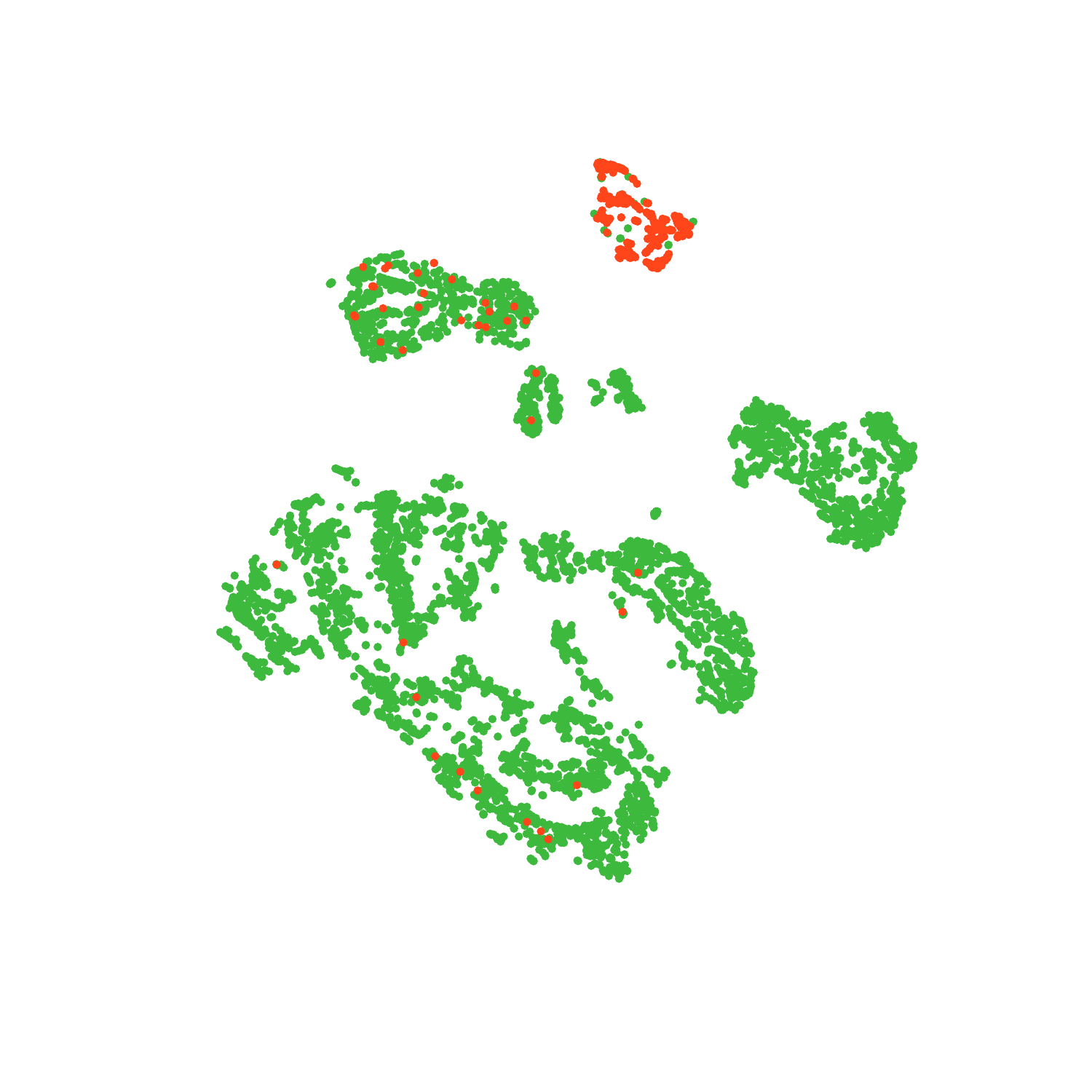}
        \caption{ConsisGAD}
        \label{fig}
    \end{subfigure}
    \begin{subfigure}{0.24\textwidth}
        \centering
        \includegraphics[width=\linewidth]{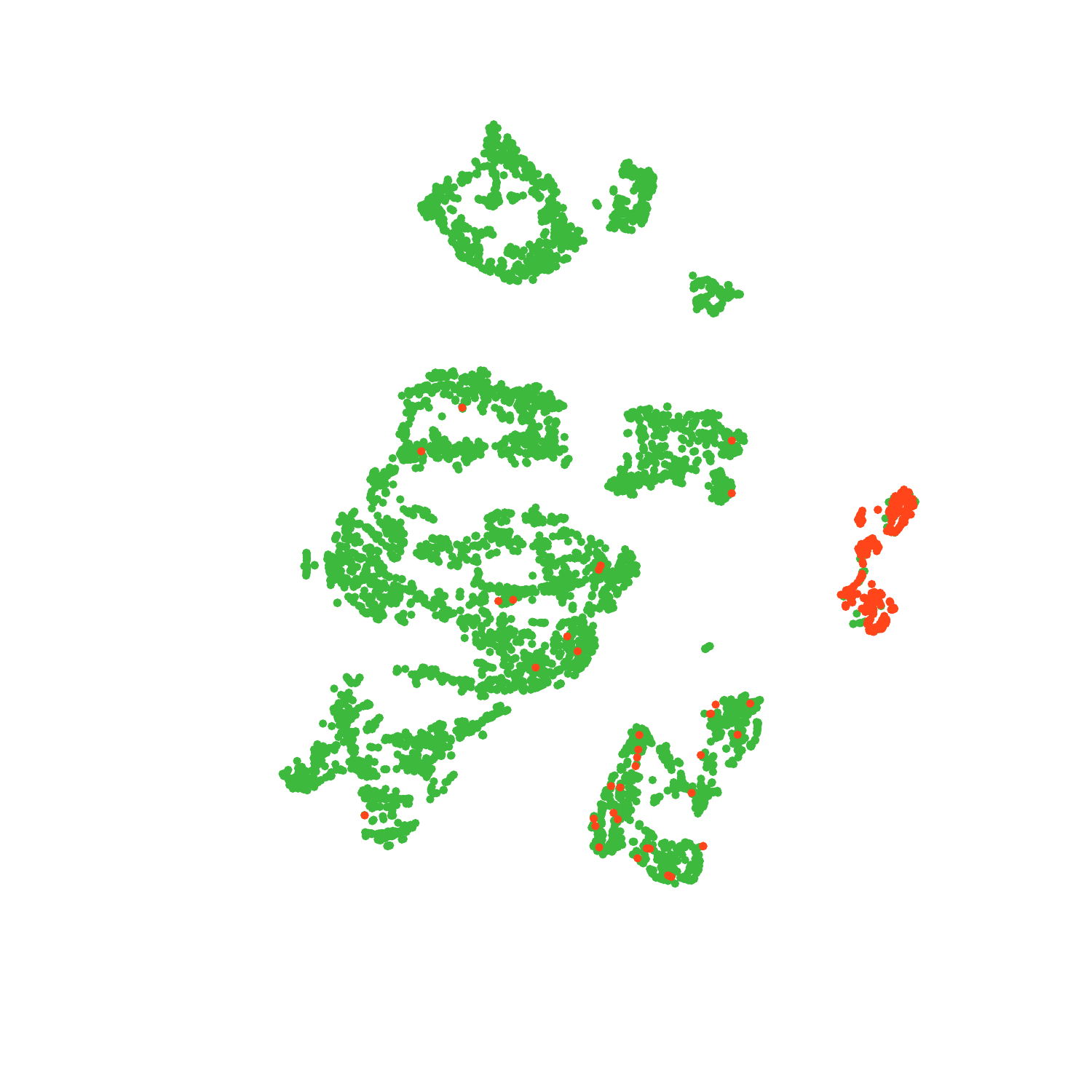}
        \caption{SpaceGNN}
        \label{fig}
    \end{subfigure}
    \begin{subfigure}{0.24\textwidth}
        \centering
        \includegraphics[width=\linewidth]{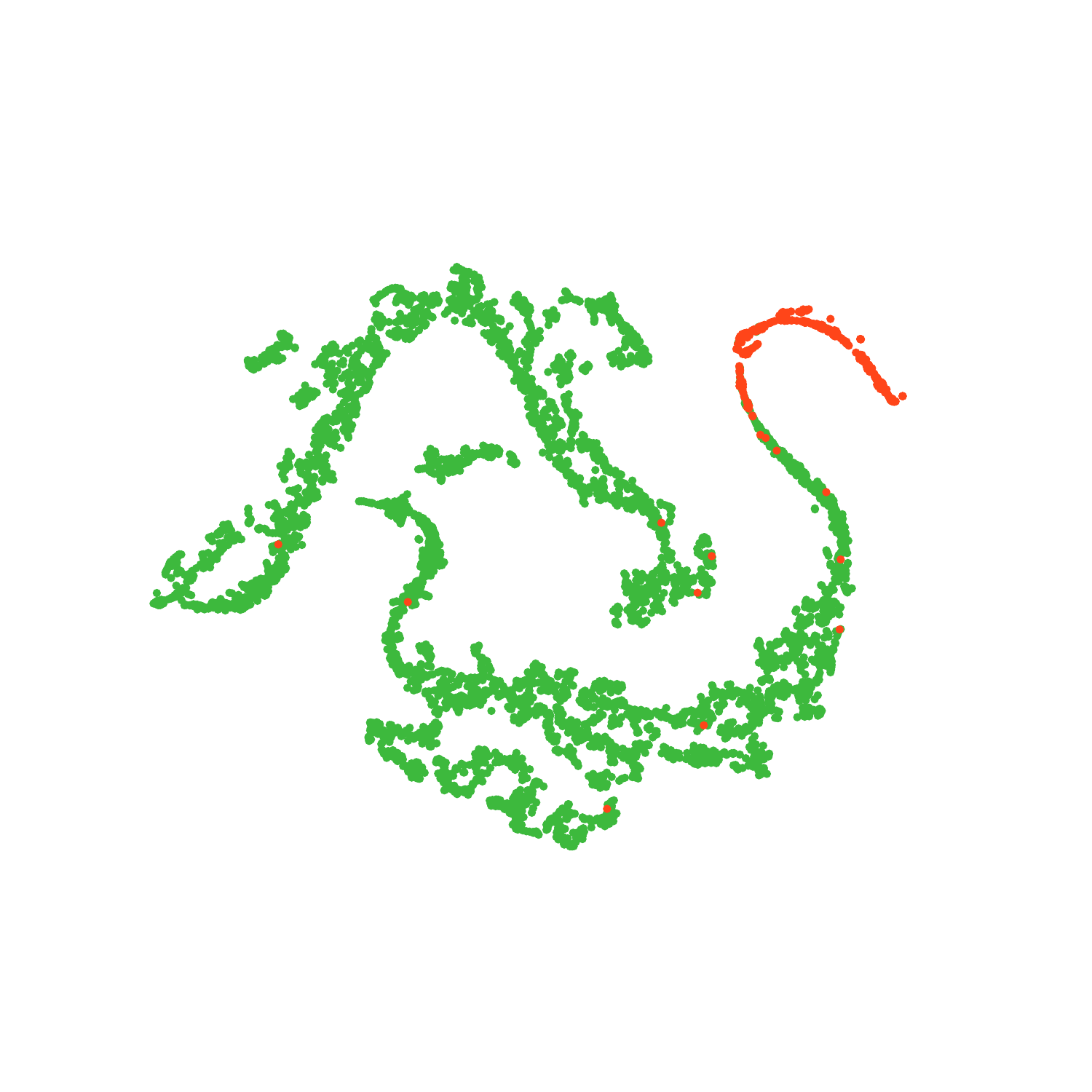}
        \caption{HIMVH}
        \label{fig}
    \end{subfigure}
  \caption{Embedding visualization of different models.}
  \label{fig:8subfigures}
\end{figure*}

As shown in Figure 5, HIMVH maintains stable performance across various hyperparameter settings, highlighting its robustness. Specifically, Figure 5 (c) shows that shallow GNNs underfit due to limited receptive fields and insufficient structural information propagation, while deeper models (4–5 layers) avoid significant degradation, attributed to CNHL, which functions as a heterophily-aware graph learning paradigm and mitigate over-smoothing encountered in deeper GNNs. Figure 5 (d) further indicates that performance improves with larger temporal window sizes, then plateaus, as small windows $w$ restrict temporal context required for modeling hyperedge center.

\subsection{Visualization and Interpretability Analysis}

To assess the effectiveness of HIMVH, we utilize t-SNE \cite{van2008visualizing} on different graph learning models. Specifically, we conduct experiments on the Private-4 dataset. Due to the extreme class imbalance, we visualize all fraud samples along with 10\% of the normal samples to ensure a clearer and more informative representation. The visualization results of eight SOTA models are presented in Figure 6, where red nodes represent online fraud behaviors and green nodes denote online normal behaviors. Although models (a)-(g) achieve commendable decoupling effects, some red nodes still appear within green regions. In contrast, HIMVH model demonstrates superior decoupling performance in distinguishing between normal and fraud nodes, as evidenced by the clear separation of red and green points in the embedding visualization. This suggests that HIMVH is more effective in capturing underlying fraud patterns and learning discriminative representations compared to existing models. 

\begin{figure}[t]
  \centering
    \includegraphics[width=0.48\textwidth]{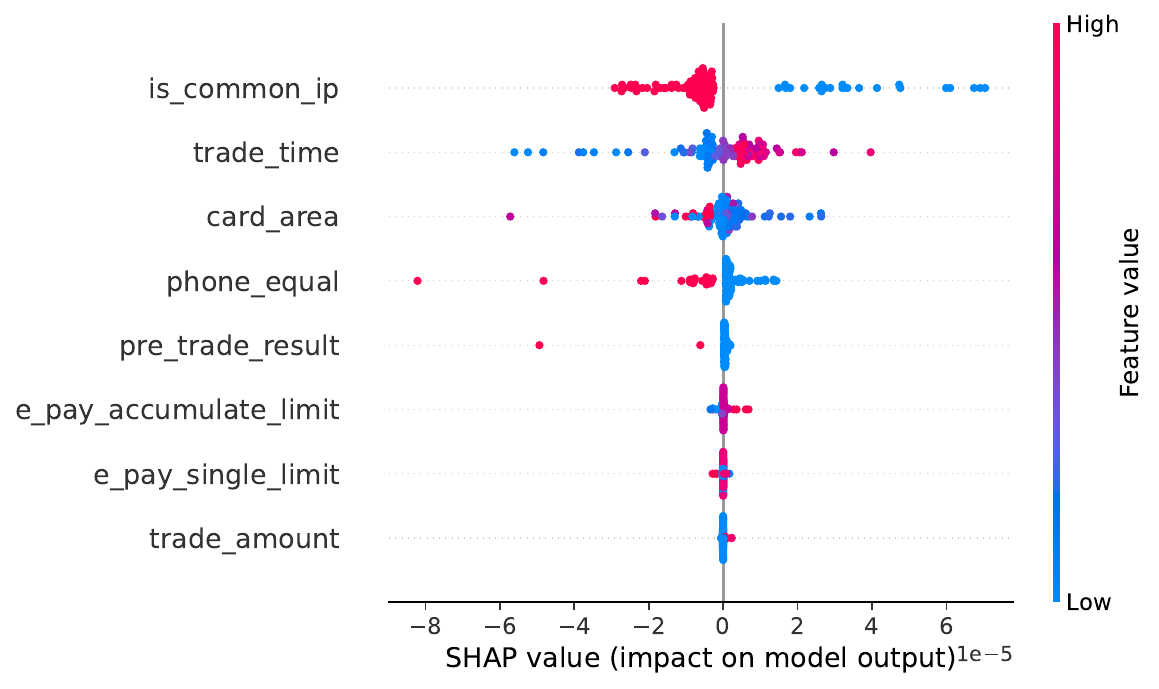}
  \caption{SHAP analysis on Private-4.}
  \label{Figure1}
\end{figure}
We further perform SHAP \cite{Lundberg2017AUA} analysis to enhance model interpretability (Figure 7). The color gradient from blue to red indicates increasing feature values, and features higher on the vertical axis contribute more to the model’s decisions. The `is\_common\_ip' feature stands out: blue dots (uncommon IPs) are associated with higher SHAP values, indicating greater fraud risk. This aligns with fraud patterns in which attackers often use ephemeral or rarely seen IP addresses to avoid being linked to known benign user behavior, thereby evading IP-based detection heuristics. Our SHAP-based analysis illustrates how the distributional discrepancies of different features influence the decision boundaries learned by HIMVH.

\subsection{Discussion}
We conduct a comparative analysis of several graph-based web finance fraud detection models and provide an explanation that improves interpretability, thereby supporting safer web financial environments and advancing broader social good.

GTAN, though equipped with temporal attention, applies a uniform aggregation and static embeddings, and fails to capture semantic inconsistencies. U-A2GAD adopts fixed kNN-based graph construction and polynomial spectral filtering, which restrict its adaptability to local structural variations. As a result, it struggles to respond to behavioral inconsistencies and camouflaged fraud patterns that deviate from the dominant transactional structure. ConsisGAD emphasizes consistency regularization under low supervision but underutilizes labeled data. This weakens its discriminative power in complex scenarios, and boundaries where fraud is sparse and ambiguous. UniGAD unifies multi-level tasks through subgraph sampling and a shared fusion module. However, its sampler prioritizes nodes with pronounced signals while discarding neighbors that appear normal yet convey essential relational cues. This bias narrows the retrieved subgraphs and limits the UniGAD’s capability to capture full anomaly contexts.

While RGTAN, and SpaceGNN perform well by modeling temporal patterns or alleviating label sparsity, they lack explicit mechanisms for identifying camouflaged fraud that mimics normal behavior and for capturing rare fraudulent activities situated at the periphery of the data distribution. These limitations reduce their effectiveness in scenarios where subtle behavioral inconsistencies and low-frequency fraud signals are critical for accurate detection.

\section{Conclusion}
In this work, we propose HIMVH, a hippocampus-inspired multi-view hypergraph learning framework tailored for web finance fraud detection. Inspired by hippocampus's scene conflict detection mechanism, we propose hippocampal cross-view discrepancy perception module. Specifically, we quantify the behavioral discrepancies of the same online transaction across different views, capturing latent cross-view conflicts from multiple dimensions to characterize camouflaged patterns more comprehensively. In addition, inspired by the match–mismatch novelty detection mechanism in the hippocampal CA1 region, we propose a heterophily-sensitive graph learning paradigm. This approach employs a deviation-aware message passing strategy, which adaptively reweights node messages based on their divergence from local neighborhood structures, thereby amplifying subtle anomaly signals from rare tail instances. Experimental results on six real-world web finance fraud datasets show that HIMVH achieves 6.42\% improvement in AUC, 9.74\% in F1 and 39.14\% in AP on average over 15 SOTA models. 

In the future, we plan to extend HIMVH to broader fraud detection scenarios beyond web finance domain, including but not limited to fake news dissemination and identity fraud.

\section{Acknowledgments}
This work is supported in part by the National Natural Science Foundation of China under Grant 62302337, 62402098, and in part by the Fundamental Research Funds for the Central Universities under Grant 2232024D-25.

\bibliographystyle{ACM-Reference-Format}
\bibliography{samples/reference}

\appendix

\section{Implementation Details of the Baselines}
\subsection{Models Without Graph Learning}

\begin{itemize}[
    label=\textbullet, 
    labelindent=0pt, 
    leftmargin=25pt, 
]
\item \textbf{MCNN \cite{fu2016credit}}. MCNN is a fraud detection model, which constructs feature matrix with transaction data. MCNN is the pioneering work that applies convolutional neural networks (CNNs) to financial fraud detection.
\item \textbf{STAN \cite{cheng2020spatio}}. STAN employs a spatial-temporal attention mechanism to model transaction records. However, since the datasets utilized in our experiments lack spatial information, we exclude the spatial component from STAN in our implementation.
\end{itemize}

\subsection{Graph Learning Models}

\begin{itemize}[
    label=\textbullet,
    labelindent=0pt,
    leftmargin=25pt,
]

\item \textbf{BWGNN \cite{tang2022rethinking}}. BWGNN captures abnormal patterns by focusing on high-frequency spectral energy shifts, offering improved detection performance.

\item \textbf{GHRN \cite{gao2023addressing}}. GHRN tackles the heterophily problem in graph by utilizing the smoothness index and graph Laplacian to identify high-frequency components. It employs node predictions to estimate high-frequency signals, and ensures robust identification even with limited labeled data.

\item \textbf{GTAN \cite{xiang2023semi}}. GTAN passes messages among nodes via a gated temporal attention mechanism to learn transaction representations, and model fraud patterns through risk propagation.
\item \textbf{U-A2GAD \cite{li2025universal}}. U-A2GAD employs enhanced high-frequency filters and integrates k-NN and k-FN graphs to effectively address low-frequency constraints and camouflage issues in GNN-based anomaly detection.

\item \textbf{ConsisGAD \cite{chen2024consistency}}. ConsisGAD addresses limited supervision and class imbalance in graph anomaly detection via consistency training and learnable data augmentation, and enhances detection performance by leveraging differences in node homophily distribution within a GNN backbone.

\item \textbf{UniGAD \cite{lin2024unigad}}. UniGAD unifies node-level, edge-level, and graph-level anomaly detection using the Maximum Rayleigh Quotient Subgraph Sampler (MRQSampler) and GraphStitch Network, improving multi-task learning performance.

\item \textbf{RGTAN \cite{xiang2025enhancing}}. RGTAN constructs a temporal transaction graph and uses a gated temporal graph attention mechanism to learn adaptive transaction representations and enhance multi-hop risk structure perception for improved credit card fraud detection.

\item \textbf{SpaceGNN \cite{dong2025spacegnn}}. SpaceGNN is a multi-space graph neural network for node anomaly detection with extremely limited labels, leveraging learnable space projection and distance-aware propagation.
\end{itemize}

To adapt the node-level anomaly detection baselines, we employ the same temporal graph construction approach as GTAN in the task of transaction fraud detection. Specifically, we construct the temporal graph by representing individual transaction records as nodes and establishing edges based on the temporal relationships between transactions.

\begin{table*}[htbp]
\centering
\caption{Fraud detection performance (\%) on six datasets when temporal window size $w$ is set to 2.}
\label{tab:results}
\setlength{\tabcolsep}{3.5pt}
\begin{tabular}{c*{7}{c@{\hspace{3pt}}c@{\hspace{3pt}}c}}
\toprule

\multirow{2}{*}{Method} & \multicolumn{3}{c}{S-FFSD} & \multicolumn{3}{c}{Sparkov} & \multicolumn{3}{c}{Private-1} & \multicolumn{3}{c}{Private-2} & \multicolumn{3}{c}{Private-3} & \multicolumn{3}{c}{Private-4} \\
\cmidrule(r){2-4} \cmidrule(lr){5-7} \cmidrule(lr){8-10} \cmidrule(lr){11-13} \cmidrule(lr){14-16} \cmidrule(lr){17-19} \cmidrule(l){20-22}
 & AUC & F1 & AP & AUC & F1 & AP & AUC & F1 & AP & AUC & F1 & AP & AUC & F1 & AP & AUC & F1 & AP \\
\midrule
HIMVH & 87.30 &	76.24 & 71.90 & 97.01 & 91.51 & 87.15 & 98.41 & 92.70 & 89.86 & 98.73 & 88.44 & 83.95 & 99.04 & 86.11 & 74.27 & 98.01 & 85.51 & 68.79\\
GTAN & 80.19 & 73.79 & 63.37 & 94.01 & 68.22 & 54.37 & 97.87 & 92.05 & 86.49 & 98.04 & 87.19 & 73.11 & 99.16 & 67.80 & 52.35 & 96.66 & 77.40 & 58.18\\
RGTAN & 79.97 & 71.45 & 63.40 & 95.58 & 81.00 & 67.01 & 97.82 & 80.16 & 87.31 & 96.81 & 86.47 & 67.46 & 97.47 & 80.57 & 55.05 & 95.70 & 80.29 & 56.22\\
SpaceGNN & 85.97 & 59.33 & 65.28 & 96.12 & 73.42 & 75.66 & 97.63 & 85.89 & 84.73 & 98.70 & 79.06 & 77.26 & 98.29 & 64.84 & 65.30 & 96.36 & 56.73 & 51.72\\
\bottomrule
\end{tabular}
\label{table3}
\end{table*}
\begin{figure*}[htbp] 
  \centering
    \begin{subfigure}{0.32\textwidth}
        \centering
        \includegraphics[width=\linewidth]{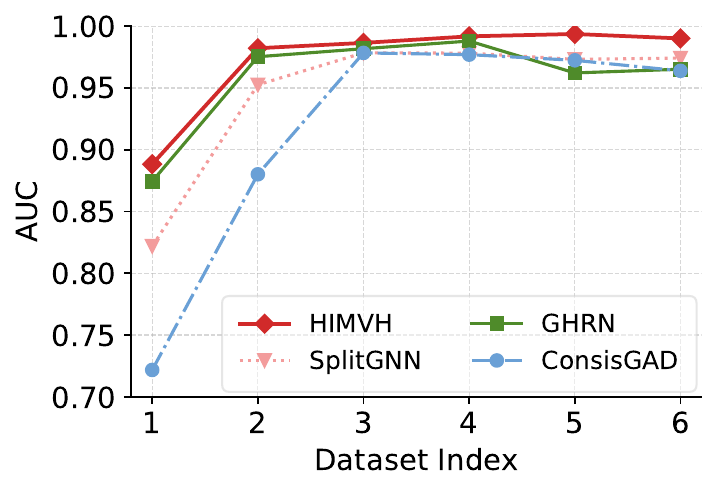}
        \label{fig}
    \end{subfigure}
    \begin{subfigure}{0.32\textwidth}
        \centering
        \includegraphics[width=\linewidth]{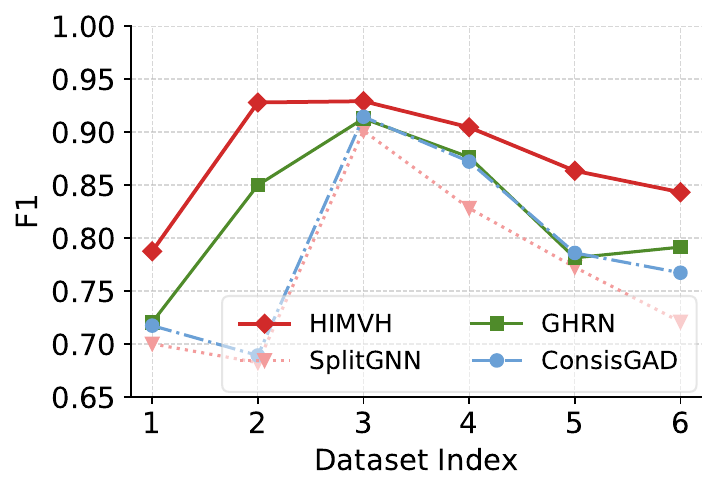}
        \label{fig}
    \end{subfigure}
    \begin{subfigure}{0.32\textwidth}
        \centering
        \includegraphics[width=\linewidth]{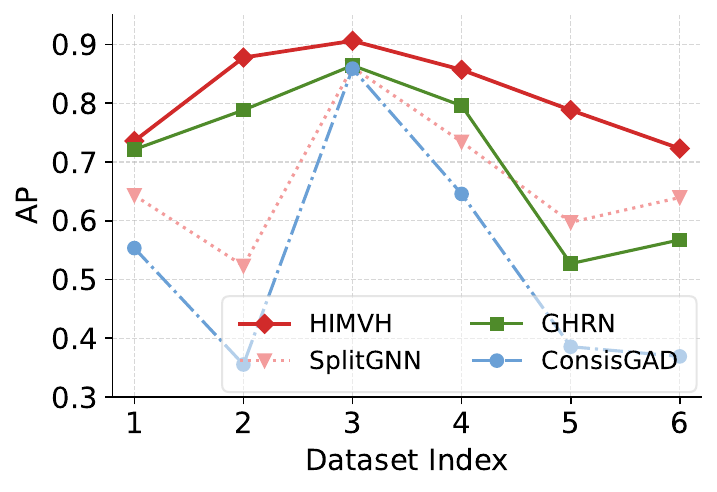}
        \label{fig}
    \end{subfigure}
  \caption{The performance of different heterophily-aware GNN models.}
  \label{fig:8subfigures}
\end{figure*}
\section{Details of Private Dataset}
The Private dataset is collected by our partner, a major commercial bank, and consists of online customer transaction records. The web-based transactions are chronologically divided into four monthly subsets, from January to April, denoted as Private-1 to Private-4. Each record is associated with one of three label types: `0' indicates a normal transaction, `1' denotes a fraudulent transaction, and `2' represents missing labels.

In addition to the identity information of the sender and receiver, each transaction contains eight attribute features, namely `trade\_time', `trade\_amount', `card\_area', `pre\_trade\_result', `phone\_\\equal', `is\_common\_ip', `e\_pay\_single\_limit', and `e\_pay\_accumulate\\\_limit'.  Detailed descriptions of these features are as follows:

\begin{itemize}[
    label=\textbullet, 
    labelindent=0pt, 
    leftmargin=25pt, 
]
\item \textbf{trade\_time}: Time of the transaction.
\item \textbf{trade\_amount}: The amount of a single transaction.
\item \textbf{card\_area}: Geographical region associated with the transaction.
\item \textbf{pre\_trade\_result}: Outcome of the previous transaction attempt.
\item \textbf{phone\_equal}: Match between transaction and registered phone.
\item \textbf{is\_common\_ip}: Whether the IP address is commonly used.
\item \textbf{e\_pay\_single\_limit}: Limit for single electronic payment.
\item \textbf{e\_pay\_accumulate\_limit}: Cumulative limit for payments over time.
\end{itemize}

\section{Additional Experimental Results}
\subsection{Long-Tail Performance}
In this section, to validate the effectiveness of the HIMVH model on the tail portion of the distribution, we set the temporal window $w$ to 2 during graph construction. This choice draws on prior insights \cite{han2025mitigating} and produces a sparse graph in which all nodes fall into the tail region. We assess the model response to tail nodes by comparing the performance drop before and after this change in $w$. A model with strong robustness to tail nodes should show a small decline in performance.

When the temporal window size $w$ is set to 2, the performance of each model is reported in Table 3. Relative to the original experimental setup, the corresponding performance drop for each model is as follows:

\begin{itemize}[
    label=\textbullet, 
    labelindent=0pt, 
    leftmargin=25pt, 
]
\item \textbf{HIMVH}: HIMVH shows an average drop of 0.80\% in AUC, 1.15\% in F1, and 2.62\% in AP across the six datasets.
\item \textbf{GTAN}: GTAN shows an average drop of 0.87\% in AUC, 6.08\% in F1, and 11.45\% in AP across the six datasets.
\item \textbf{RGTAN}: RGTAN shows an average drop of 1.69\% in AUC, 3.59\% in F1, and 10.12\% in AP across the six datasets.
\item \textbf{SpaceGNN}: SpaceGNN shows an average drop of 0.64\% in AUC, 16.78\% in F1, and 6.99\% in AP across the six datasets.
\end{itemize}

Overall, HIMVH records the smallest performance drop among all models.  This outcome indicates that HIMVH maintains stable accuracy when all nodes lie in the tail region and sustains reliable results under sparse temporal structures. The consistent performance across six datasets also shows that HIMVH can extract useful signals from nodes with rare patterns and low support. These results confirm the strong capability of HIMVH in tail node analysis, with rare or subtle fraud behaviors.

\subsection{Heterophily Analysis}
A heterophily-aware GNN refers to a graph model that can capture node relations in graphs where connected nodes often hold different labels or attributes. As a heterophily-aware GNN, HIMVH assigns higher weight to node pairs with larger similarity difference. We compare HIMVH with three heterophily-aware models, SplitGNN \cite{xu2023splitgnn}, GHRN \cite{gao2023addressing}, and ConsisGAD \cite{chen2024consistency}, where SplitGNN is not part of the 15 SOTA baselines. Figure 8 reports the results on the six datasets in index order from one to six. HIMVH has an average gain of 2.72\% in AUC, 14.35\% in F1, and 22.26\% in AP over SplitGNN. The results confirm the clear advantage of HIMVH over other heterophily-aware GNN models.

\subsection{Time Efficiency}

\begin{table}[h]
\renewcommand\arraystretch{1.0}
\centering
\caption{Inference time (s) of different models.}
\begin{tabular}{ccc ccc | ccc ccc}

\toprule[1pt] 
\multicolumn{3}{c}{Dataset} & \multicolumn{3}{c}{Time}  &  \multicolumn{3}{c}{Dataset} & \multicolumn{3}{c}{Time}   \\

\toprule[1pt] 

\multicolumn{3}{c}{GTAN} & \multicolumn{3}{c}{1.44} & \multicolumn{3}{c}{RGTAN} & \multicolumn{3}{c}{1.79} \\

\multicolumn{3}{c}{HIMVH} & \multicolumn{3}{c}{2.09} & \multicolumn{3}{c}{SpaceGNN} & \multicolumn{3}{c}{3.42} \\

\bottomrule[1pt] 
\end{tabular}
\end{table}

Under the same setup as in the main paper, we assess the inference time on the Private-4 dataset across twenty runs. The inference results appear in Table 4, GTAN costs 1.44 s, RGTAN costs 1.79 s, HIMVH costs 2.09 s, and SpaceGNN costs 3.42 s. These results show that our method secures clearly stronger detection performance while its computational cost remains fully acceptable. The time reported for HIMVH reflects a single forward pass within one view, which indicates that even with its richer feature use and higher accuracy, the overall expense stays well within a practical range for real fraud analysis scenarios.

\end{document}